# ASSESSING OF SOIL EROSION RISK THROUGH GEOINFORMATION SCIENCES AND REMOTE SENSING - A REVIEW

Lachezar Filchev[1], Vasil Kolev[2]

**Abstract:** During past decades a marked manifestation of widespread erosion phenomena was studied worldwide. Global conservation community has launched campaigns at local, regional and continental level in developing countries for preservation of soil resources in order not only to stop or mitigate human impact on nature but also to improve life in rural areas introducing new approaches for soil cultivation. After the adoption of Sustainable Development Goals of UNs and launching several world initiatives such as the Land Degradation Neutrality (LDN) the world came to realise the very importance of the soil resources on which the biosphere relies on for its existence. The main goal of the chapter is to review of different types and structures erosion models as well as its some applications. Several methods using spatial analysis capabilities of geographic information systems (GIS) are in operation for soil erosion risk assessment, such as: Universal Soil Loss Equation (USLE), Revised Universal Soil Loss Equation (RUSLE) in operation worldwide and in USA and Modèle d'Evaluation Spatiale de l'ALéa Erosion des Sols (MESALES) model. These and more models are being discussed in present work alongside more experimental models and methods for assessing soil erosion risk such as Artificial Intelligence (AI), Machine and Deep Learning etc. At the end of this work a prospectus for the future development of soil erosion risk assessment is drawn.

**Keywords:** Soil erosion; assessment risk, Universal Soil Loss Equation (USLE), Revised Universal Soil Loss Equation (RUSLE), RUSLE2 (Revised Universal Soil Loss Equation, version 2.0), machine learning (ML), Geographical Information systems (GIS), Artificial neural networks (ANN), Convolutional neural network (CNN)

[1]Lachezar Filchev is with Space Research and Technology Institute, Bulgarian Academy of Sciences, Bl.1 Acad. G. Bonchev St., 1113 Sofia, Bulgaria, E-mail: lachezarhf@space.bas.bg
[2] Vasil Kolev is with Institute of Information and Communication Technologies, Bulgarian Academy of Sciences, Bl. 2 Acad. G. Bonchev St., 1113 Sofia, Bulgaria, E-mail: kolev_acad@abv.bg

# 1. INTRODUCTION

Soil erosion by water was found to be a ubiquitous site problem. The study of soil erosion is having become a severe worldwide problem. Moreover, it is a serious environmental, economic, and social problem as often met in agriculture. These are very real and at times severe issues. An effective soil erosion prediction has become an essential goal in many researchers. Soil erosion is considered a major threat to food security and causes damages both on-site and off-site, i.e., to adjacent infrastructures and surface waters. Many soil properties used for design are not intrinsic to the soil type, but vary depending on conditions. In-situ stresses, changes in stresses, the presence of water, rate and direction of loading, and time can all affect the behavior of soils.

New challenges and policy developments after 2015 (among others, the Common Agricultural Policy (CAP), Sustainable Development Goals (SDGs)) are opportunities for soil scientists and soil erosion modellers to respond with more accurate assessments and solutions as to how to reduce soil erosion and furthermore, how to reach Zero Net Land Degradation targets by 2030. According to the Thematic Strategy for the Soil Protection in European Union (EU) (European Commission - Soil Thematic Strategy, 2006), soil erosion by water is the most severe hazard for soils in Europe. In 2019 five mission areas centered on societal priorities and challenges have been approved in EU. One of them 'Soil health and food' is aiming at least 75% of all soils in each EU Member States 2030 to turn healthy, or show a significant measurable improvement for all indicators where levels are below accepted thresholds, to support the provisioning of essential ecosystem services (Veerman et al. 2020). In order to estimate this Mission defining a set of 8 fundamental indicators the fourth of which is '4) Soil structure including soil bulk density and absence of soil sealing and erosion'. This fundamental indicator is to be observed and estimated EU wide by the EU Soil Observatory which in turn is run by the DG JRC of EC.

When estimating the *soil erosion risk*, it is recommended to exclude surfaces that are not prone to soil erosion such as urban areas, bare rocks, glaciers, wetlands, lakes, rivers, inland waters and marine waters. Apart from the rainfall and topographic factors, soil properties and cover management conditions influence the variation of soil loss rates. However, there are big variations in rainfall distribution and slope gradient, which lead to high fluctuations in the soil erosion risk from one place to another. The rate of the potential *erosion risk* is increasing with the increase of the rainfall intensity, slope gradient, and soil erodibility. Soil erosion prediction is relevant at a wide range of spatial scales, from the plot scale to the catchment scale, from the regional scale up to the continental and global scales. At different scales, different processes tend to become dominant, so that the effective focus of the models also changes. At the larger scales, topography, soil and vegetation patterns become more important. This is where remote sensing and GIS become valuable tools.

*Remote sensing* has become a monitoring and predicting tool for environmental variables by using satellite data which have clear benefits for and can be delivered information as fast as changing surface conditions. Remote sensing imagery can be used in soil erosion modeling where is studies land degradation of erosion features such as gullies or vegetation impoverishment or by collect input data for simulation. Methods using satellite imagery to produce maps of vegetation-related variables for soil erosion studies have been compared, who found that the normalized difference vegetation index (NDVI) was the most useful.

*Geographical Information systems* (GIS) provide an important spatial and analytical function, for geo- and spatial processing to develop the erosion model. Some important advantages include in soil erosion models are:
- Processing of the input data to simulate different scenarios very quickly.
- The ability to look at spatial variation and applied of a custom resolution;
- Use of an animate sequence of model output across time and space. It is enables objects to be viewed from all external perspectives.

The use of GIS is indispensable in agroforestry research, planning and extension because agroforestry science is directly dependent upon the linkage of spatial and non-spatial information, for instance use of spatially-distributed soil erosion models with GIS that can quantify soil loss at the landscape level and spatial erosion risk assessment can be determined. Also the GIS techniques can provide easy and time effective tools to map and analyze erosion input data of hydrophysical parameters. The well-know and often applicable models in worldwide are USLE and RUSLE. Because of they have compatibility with GIS (Millward and Mersey, 1999; Jain et al. 2001; Pandey et al. 2009; Lu et al. 2004; Jasrotia and Singh, 2006; Dabral et al. 2008; Kouli et al. 2009; Bonilla et al. 2010; Samanta et al. 2016)). In 1993 using RUSLE begun the development of the RUSLE2 model as uses the basic USLE equation structure to compute sediment detachment but differs greatly from the USLE in almost every other way. RUSLE 2 is similar to RUSLE, but RUSLE 2 uses new equations, a new mathematical integration procedure, new database values, and is implemented in a modern graphical user interface computer program. Almost all of the mathematical relationships in RUSLE2 have been revised from corresponding relationships in RUSLE.

The main goal of soil erosion models is either the predictability or explanatory. Modeling of soil erosion is the process of describing soil particle detachment, transport and deposition mathematically on land surfaces. It is also necessary to keep in mind that good predictions can only be obtained by using models if they are based on good data from field observations and measurements, and not on generated or assumed data. Soil erosion models are usually developed for particular environments, which mean they only involve erosion processes that are important in those environments based on various assumptions and simplifications. Soil erosion models have been widely applied to predict soil erosion rates and understand interactions between different environmental drivers of soil erosion processes.

In last decade years, many erosion models are appeared includes some extension and improvements of already existing methods, new methods based of AI, as well as erosion models dependents from geography areas are constructed. There exist different classifies of the erosion models – mains, spatials, evens types, scale types, categories, structures, Artificial Intelligence (AI) – based, ML-based, and so on, which leads to appears of the big important question – Is it possible let to do a new, a global division of erosion models ? It is easy to see that such a division can be – *Human Intelligence* (HI) and AI - *based* erosion models. It is allows us also easily add and division of further erosion models, which most likely should will be the Artificial Intelligence – based models.

Therefore, the review chapter considers most-used erosion models divided of two categories which are described in the following parts, respectively. In the 'Soil erosion' is considered in brief the erosion theory, some most important characteristics and divisions. In last subchapter are discussed some of the models and conclusions are drawn.

## 2. SOIL EROSION

### 2.1 General Theory

Soil erosion is the common land degradation problem in the worldwide because of its economic use and environmental impacts. To estimate soil erosion and to establish soil erosion management plans, many computer models have been developed and used.

*Soil erosion* is a complex geomorphological process with varying influences of different impacts at different spatio-temporal scales. Soil erosion is the displacement of the upper layer of soil, one form of soil degradation. This natural process is caused by the dynamic activity of erosive agents, that is, water, ice (glaciers), snow, air (wind), plants, animals, and human activities. In accordance with these agents, erosion is sometimes divided into water erosion, glacial erosion, snow erosion, Soil erosion may be a slow process that continues relatively unnoticed, or it may occur at an alarming rate causing a serious loss of topsoil. The loss of soil from farmland may be reflected in reduced crop production potential, lower surface water quality and damaged drainage networks. The assessment methods of the soil erosion processes at field scale and under natural conditions is necessary to measure soil surface changes for multi-spatio-temporal scales without disturbing the area of interest and to identifies and quantifies sediment sources and sinks at the hillslope with high spatial resolution.

### 2.2 Erosion Types

The development of soil properties for geotechnical design purposes begins with developing the geologic strata present. A geologic stratum is characterized as having the same geologic depositional history, stress history, and degree of disturbance, and generally has similarities throughout the stratum in terms of density, source material, stress history, hydrogeology, and macrostructure. The properties of each stratum shall be consistent with the stratum's geologic depositional and stress history, and macrostructure

The main types of erosion studied by the community are:

- ➢ *Impact Erosion* – Physical detachment of soil particles as a result of raindrop impact.
- ➢ *Sheet erosion* – Thin, uniform wearing away of the uppermost surface layers in the soil profile. Seldom have the detaching agent, but just merely transporting soiled particles detached by raindrop impact.
- ➢ *Rill erosion* – Follows sheet erosion. As the amount and velocity of water increases water is now able to both detach and transport soil particles.
- ➢ *Gully erosion* – As rills deepen and widen, gullies form. Simple definition: gullies are rills that are too large to be repaired with conventional tillage equipment. Gullying is a type of *soil erosion* that currently represents a major threat at the societal scale and will likely increase in the future.
- ➢ *Channel erosion* – Erosion as a result of concentrating and confining the erosive forces of water. Includes both manmade and natural channels.
- ➢ *Mass Wasting* – Large failures usually as a result of gravitational forces. Landslides, pot-slides, slumps, debris torrents

- *Soil erosion* is a complex geomorphological process with varying influences of different impacts at different spatio-temporal scales, the displacement of the upper layer of soil, one form of soil degradation. This natural process is caused by the dynamic activity of erosive agents, that is, water, ice (glaciers), snow, air (wind), plants, animals, and human activities.

**2.3 Erosion Model Types and Structures**

Models are part of the erosion prediction technology, used in development of soil conservation and environmental degradation planning systems. Erosion is based on defining the relation among controlling factors and soil loss delivery soil loss from one point on the landscape to another within a given set of management conditions.

The types of soil erosion models are summarized in (Wischmeier and Smith, 1978). More recently an overview with a typologization of water erosion models is made by (Karydas et al., 2012). In general, models fall into three main categories, depending on the physical processes simulated, the model algorithms describing these processes, and the data dependency of the model. The category of soil erosion models:

- *Empirical models* – They are generally refers to a simplified representation of a system and are based primarily on the analysis of field experiments and seek to characterize the response from these erosion plots using statistical inference. The computational and data requirements for such models are usually less than for conceptual and physically based models (SLEMSA, MUSLE, USLE, RUSLE etc.). They are particularly useful as a first step in identifying the sources of sediments.

- *Conceptual models* (Beck, 1987; Renschler, 1996; Merritt et al., 2003) – include only a general description of catchment processes, without including the details occurring in the complex process of interactions. This allows these models to provide an indication of the qualitative and quantitative effects of land use changes, without requiring large amount of spatially and temporally distributed input data.

- *Physically-based models* – They are based on an understanding of the physics of the erosion and sediment transport processes and describe the sediment system as includes the laws of conservation of mass and energy, where energy can change form but total energy remains the same. They are based on the understanding of the physics of erosion processes. Example models are Soil and Water Assessment Tool (SWAT) (Arnold and Fohrer, 2005), WEPP, Chemicals, Runoff and Erosion from Agricultural Management (CREAMS) (Knisel, 1985), Aerial Non-Point Source Watershed Environment Response Simulation (ANSWERS) (Beasley et al., 1980), The Three Dimensional Storm Process Based Erosion model (EUROSEM) (Morgan et al., 1998), and Agricultural Non-Point Source Pollution model (AGNPS) (Young et al., 1989; Bhuyan et al., 2001; Grunwald and Norton, 1999), etc. The equations are formulated for use with continuous spatial and temporal data, yet the data used in practice are often point source data to represent. The main advantage of physical models is more their relative transferability, which favours predicting soil erosion and sediment yield under different climate and physiographic land use scenarios and their ability to consider environmental issues such as climate change.
Another typologization separates them into two types of soil erosion models:

- *Empirical* – In these models, statistical techniques are employed to examine the relationships between different components of studied systems. Empirical models can achieve accurate results. However, they are limited to conditions for model development and therefore often do not perform well when applied to other areas or other time periods
- *Process-based* – They explain and predict the dynamic behaviour of the system.

We also have **two structures** of soil erosion models:

- *Lumped* – In lumped models, contributing factors of erosion are represented by a constant value over the study area.
- *Distributed* – In spatially-distributed models, by using of advances of GIS the spatial variability achieve better representations, a possibility of a large area to be divided into small sub-units which have uniform characteristics - climate, land use, and topography.

Considering the spatial and temporal domain – **two scales** of a soil erosion model focuses on soil erosion and transport at:

- *Spatial scale* – It refers to the spatial extent or time span that models operate at;
    1. Hillslope ($< 0.1$ km$^2$),
    2. Small catchment ($0.1$ km$^2$–$1000$ km$^2$),
    3. Medium catchment ($1000$ km$^2$–$10,000$ km$^2$),
    4. Large catchment/regional ($> 10,000$ km$^2$),

- *Time step* – The time interval used by a model during applications:
    1. Event,
    2. Daily,
    3. Monthly,
    4. Annual.

The models can be classified also according to:

- The spatial as:
    1. *Non-spatially distributed* – EPIC, CREAMS,
    2. *Spatially distributed* – ANSWERS, AGNPS, and SWRRB.

- The evens type as:
    1. *Single-event* – AGNPS, ANSWERS,
    2. *Continuous-time scale* – EPIC, CREAMS, and SWRRB.

> The scale type as:

1. *Field-scale* – WEPP, EPIC, CREAMS,
2. *Watershed/basin-wide* – ANSWERS, AGNPS, and SWRRB.

The primary factors for soil erosion are identified as: *climate*, *soil*, *topography*, *vegetation,* and *conservation* practices, which have either a positive or negative effect on soil loss. These primary factors are the basis of the universal soil loss equation (USLE) model.

**2.4 Some Important Erosion Model Parameters**

Considering the soil erosion model's parametrization with remotely sensed data and products the most widely used vegetation index is the *Normalized Difference Vegetation Index* (NDVI). NDVI is one of the most relevant vegetation indices used heavily in different disciplines to estimate vegetation cover with sufficient accuracy. Many researchers used NDVI as a reliable method for geospatially estimating vegetation status.

The NDVI (Karaburun, 2010) express the difference between reflectance in the Red and Near-Infra-Red (NIR) portion of the electromagnetic spectrum, and by extension between the corresponding satellite bands and used in large geographic areas. For LANDSAT 8 Red portion is located in Band 4 and Near-Infra-Red (NIR) in Band 5. Therefore the index is definite from the following formula (Efthimiou and Psomiadis, 2018):

$$\text{NDVI} = \frac{\text{NIR} - \text{Red}}{\text{NIR} + \text{Red}} = \frac{\text{Band5} - \text{Band4}}{\text{Band5} + \text{Band4}}, \quad -1 \leq \text{NDVI} \leq 1$$

The index values are in the intervals:
  NDVI < -0.2 – water bodies,
  NDVI < 0.2 – the non-vegetated areas (bare soil, rock, snow, built-up areas),
  NDVI > +0.2 – vegetation traces,
  NDVI = [0.2–0.4] – represent shrub and grassland,
  NDVI = [0.6–0.9] – indicate temperate and tropical rainforest.

The index has strong correlations with Leaf Area Index (LAI), total vegetation cover, or above-ground biomass and can be used to express the vegetative state in different seasons to find a soil erosion risk assessment. To obtain of the NDVI images can be acquired by the spectroradiometer sensor on board the TERRA satellite (Alexandridis et al. 2015) and can be used to provide a good estimate of the annual soil loss. Since finding of the soil loss of each month is compared with the mean monthly, then the month with the lowest difference was selected as the optimum single time which can be use in a soil erosion model.

The important factor closely related the NDVI index is the coefficient C.

$$C = e^{-\alpha \frac{\text{NDVI}}{\beta \cdot \text{NDVI}}}, \quad 0 \leq C \leq 1$$

where C – the cover and management factor, $\alpha$ and $\beta$ are the parameters that determine the shape of the NDVI curve as some researches to obtain reasonable results $\alpha = 2, \beta = 1$. C-factor is related to the ratio of soil loss to soil loss occurring in bare soil. If the land use completely prevents erosion – C becomes 0; if there is no land use that acts as a protection against erosion, then the C becomes 1, and for the otherwise C < 1.

**2.5 Basic Methods, Tools, and Standards for Soil Erosion Risk Assessment**

The selection of method of measurement depends on the type of erosion, the purpose, and the target accuracy of the measurement. The techniques for measuring erosion discussed in (Loughran, 1989) are all appropriate and necessary for assessing soil loss.

## 3. EXPERT – BASED METHODS

An expert-based approach of the soil erosion risk map of Western Europe is described in (De Ploey, 1989) where the map produced by various experts who delineated areas where, according to their judgment, erosion processes are important. But, there exist that the limitation the expert – based approach does not give us the clearly for the plan criteria. Despite that expert-based method is a qualitative and its results depend heavily on expert judgment, it can be a good way to derive information about land degradation phenomena for the scale of the produced maps, as well as by lack of suitable quantitative data (Colonna et al., 2008).

**3.1 MESALES (**Modèle d'Evaluation Spatiale de l'ALéa Erosion des Sols**) Model,** or **RMSER (**Regional Modelling of Soil Erosion Risk**) Model**

The MESALES (RMSER) is based on expert-defined empirical rules and available data for the assessment of soil erosion risk at the European scale with 25 ha minimum mapping unit for land cover and 5 ha for land cover change every 6 years. (Bissonnais et al. 2002). CORINE land cover mapping provides the only consistent classification system of long-term land cover data in Europe defined in 44 classes. It is the assumption that soil erosion occurs when water that cannot infiltrate into the soil becomes surface runoff and moves downslope; thus the erosion process that is considered is erosion by overland flow. Land cover and crust formation on cultivated soils were considered as key factors influencing runoff and erosion. The soil erosion assessment in 5 classes (very low, low, moderate, high, very high) based on hierarchical decision tree (DT) classifications which are simple, requires ranking of each parameter, do not require the use of parameters that are not available at national scale, such as the USLE model, but is give priority to the human activities factors.

In compare with the CORINE erosion model the present MESALES model is much more precise and accurate uses a single decision tree whatever the land use and takes into account only two classes of land use, three climate classes and four slope classes.

The main disadvantages of the French model assessing erosion risk that the final information is provided on a 5 scales of risk and it is not possible to link these classes to quantitative values of erosion, nor is it possible to assess the errors associated with the results. The advantage of the MESALES model that allows realization of a single homogeneous map of erosion risk at a national scale, shows the importance of the seasonal effect on erosion, and gives us a possibility to be compared regions. Therefore, the decision tree MESALES model is applicable for large areas with expert opinion as it cam be use to either simulate soil erosion sensitivity or soil erosion risk.

In (Hessel et al. 2014) is made an application of the present model for 3 large areas in Europe and Morocco, using soil data from ESDB and DSMW. The newly e-SOTER database (Michéli et al. 2011) was used to evaluate whether its assessment of soil erosion sensitivity is better than existing data. The e-SOTER database consists of a spatial unit map of soil and terrain spatial units in combination with a database that gives representative soil profiles and values of soil attributes for these profiles for the units on the map. As the soil map does not contain data for these areas, MESALES is unable to generate a result as well as for application to Europe, as observed on the maps presented in JRC. Due to inconclusive expert judgment and the fact that the MESALES predicts soil erosion sensitivity, which cannot be measured in the field, in (Michéli et al., 2011) is didn't say which results are better. In (Colmar et al., 2010) MESALES and PESERA models are used and compared to obtain of the erosion risk to evaluation map at the national scale and to call upon expert knowledge to validate, at the regional scale of Brittany. The results show that both methods improved the results and we were able to produce a new regional map of erosion risk for Brittany. In (Cheviron et al., 2011) are consider and compared 4 erosion models where is shows that the PESERA model exhibited amore dynamic behavior than the MESALES through stronger variations in model outputs and sensitivity indices. MESALES behaviors is consistent with current knowledge on soil erosion as it be used to predict seasonal erosion rates on larger territories, provided sufficiently reliable input parameter values are available. Need to note, expected results are strongly affected by uncertainties in erodibility arising from imprecise cartography or lack of information on soil characteristics.

## 4. MODEL BASED METHODS

A wide variety of models are available for assessing soil erosion risk. Erosion models can be classified in a number of ways. One may make a subdivision based on the time scale for which a model can be used: some models are designed to predict long-term annual soil losses, while others predict single storm losses (event-based). Alternatively, a distinction can be made between lumped models that predict erosion at a single point, and spatially distributed models. Another useful division is the one between *empirical* and *physically-based models*.

### 4.1 Empirical models

#### 4.1.1 CORINE (Coordinated Information on the Environment) **Method**

CORINE model (CORINE, 1992) is an empirical and an expert-based factor model having simple structure predicting soil erosion. It is spatial-explicit correctly identifying Mediterranean areas with the highest risk of erosion. It is a semi-qualitative cartographic method that involves designing and overlaying of several layers-thematic maps, and it can present the spatial heterogeneity of soil erosion risk with GIS by four parameters:

- Soil erodibility;
- Erosivity;
- Slope;
- Surface cover as essential databases;

Moreover, together with GIS and remote sensing the model can be use to obtaining better accuracy in larger areas of a soil erosion risk and its spatial distribution. It was mostly applied in the European and Mediterranean countries, while spatial comparison of actual soil erosion risks map and field investigation. In (Zhu, 2012) is described that the model consists from the four steps:

**Step 1:** Soil texture, depth and stoniness layers are overlaid to form a soil erodibility map.

**Step 2:** Use of the two aridity indices to forming the erosivity layer:

- The Fournier index is $FI = \sum_{i=1}^{12} \frac{Pi^2}{P}$, where $Pi$ is total precipitation in a month, and $P$ - total mean annual precipitation;

- Bagnouls - Gaussen index $BGI = \sum_{i=1}^{12}(2t_i - P_i)k_i$

   where $t_i$ is the mean temperature for the month, $k_i$ is the proportion of the month during which $2t_i - P_i > 0$;

**Step 3:** Slope classes from digital elevation model (DEM) of the study area are produced. DEM is a valuable tool for the topographic parameterization.

**Step 4:** Prepare of the Landsat 5 TM imagery to obtain of the land use and land cover (LULC) layer.

In order to obtain the final soil erosion risk the combination between LULC and layers with the potential soil erosion risk is considered. In a study case the Dan-jiangkou Reservoir region (DRR), China, considered in (Zhu, 2012), is found that the CORINE model possess the two disadvantages, the first that it provide only qualitative output, which can be hard to validated, and second one that the final results have a smoothing effect that may erase some small or sparse high risk areas. In (Tayebi et al., 2016) are considered areas within the Bonrod Zangane watershed, western Shiraz, Iran, to assess soil erosion risk for restoring and protecting. The high soil erosion risk is found at the northern and south eastern parts of the study area. The results show that CORINE model with GIS as provides the most important factors for erosion risks in the watershed it can be used to delineate the soil erosion risk and also to discriminate the potential soil erosion risk areas.

The CORINE model is used in (Dengz and Akgul, 2005) to find of the soil erosion risk in Gölbasi Environmental Protection area and its vicinity, located south of the city of Ankara, Turkey. The model combining 4 parameters, soil erodibility, erosivity, topography and vegetation cover consists of 6 steps each of which using different overlaying combinations of soil texture, depth, stoniness, climatic data, and LULC information. The final results show that the CORINE model is very useful, in contrast the conventional methods which require high labour cost, time to collect data.

**4.1.2 USLE** (Universal Soil Loss Equation) **Model**

The USLE is an empirical model was designed by the United States Department of Agriculture (USDA) in 1978 to predict longtime-average inter-rill and rill cropland soil losses by water under various effects such as land use, relief, soil and climate, and guide development of conservation plans to control erosion. Although it is an empirical model, it predicts erosion rates of ungauged watersheds as well as presents the spatial heterogeneity of soil erosion. The model was based on the field measurements of soil erosion rates in agricultural areas in (Wischmeier and Smith, 1978). The use of USLE model is highly acknowledged in both agricultural and hilly watersheds because of its beneficial points as compared to the others tool to desegregate disparate data sets and in a position to evaluate any dynamic system like soil erosion. It is the primarily favored analytically based model used worldwide for soil erosion prediction and management, also the USLE can predict the long-term average annual soil loss associated with sheet and rill erosions. The USLE incorporates improvements in the factors based on new and better data but keeps the basis of the USLE equation. Additional research and experience have resulted in an upgrade of the USLE from the past 30 years. The USLE was enhanced by revising the weather factor, the soil erodibility factor depending on seasons, revising the gradient and length of slope and developing a new method to calculate the cover management factor. The USLE assumes that detachment and deposition are controlled by the sediment content of the flow. Erosion is limited by the carrying capacity of the flow but is not source limited. Detachment will no longer take place when the sediment load has reached the carrying capacity of the flow. USLE analysis includes R factor, K factor, LS factor, C factor and P factor values as LS factor play very important role in model. The formula of USLE is (Universal Soil Loss Equation, 2021):

$$A = R * K * L * S * C * P,$$

where,
    A is computed soil loss,
    R is the rainfall-runoff erosivity factor,
    K is a soil erodibility factor,
    L is the slope length factor,
    S is the slope steepness factor,
    C is a cover management factor,
    P is a supporting practices factor.

In (Devatha et al., 2015) using USLE model and Remote Sensing (RS) and Geographic Information System (GIS) methods are used for estimate the annual soil loss for Kulhan watershed of Shivnath basin, sub-basin of Mahanadi basin, Chhattisgarh. Results shows that the study area has gentle slope so the erosion loss is obtained with low rate and it is within acceptable limit. Soil erosion map is reclassified according to erosion risk where in the study area is within the acceptable limit. Results shows that soil erosion

USLE model in combination with GIS is an efficient tool to handle large volume data needed for watershed soil loss studies.

When the assessment model region is large and needed consider big year time period USLE method can be time-consuming. A help to reduce the calculation time significantly come by the parallelization implementation of method. Such an implementation in (Wieland et al., 2012) is considered. In (Risse et al., 1993) shows that the USLE method overestimates erosion for low measured erosion and does not underestimate erosion for measured high erosion relative to moderate erosion.

By using of USLE model and remote sensing technologies for the northern catchment of Lake Tana in (Balabathina et al., 2016) soil erosion risk showed the high influence of climatic seasonality and was high during the rainy season only. Barren land exhibited the highest soil erosion rates, followed by the croplands and plantation forest in the catchment. The soil erosion estimation was generated by multiplying the required input thematic layers of the model together in a GIS platform. The results showed that gully erosion was constrained in the steep slopes of all sub-catchment areas, which could be attributed to higher steep slopes in land-use. But also it expanded significantly to the middle and lower parts of the catchment.

**4.1.3 MUSLE** (Modified Universal Soil Loss Equation)

In (Williams, 1975) is an empirical model was developed the following revised form of the USLE using 778 storm-runoff events collected from 18 small watersheds, with areas varying from 15 to 1500 ha, slopes from 0.9 to 5.9% and slope lengths of 78.64 to 173.74 m and called it the modified universal soil loss equation (MUSLE). The MUSLE was given in the following general form:

$$S_y = a(Q \cdot q_p)^b * K * L * S * C * P$$

where,

$S_y$ is sediment yield (in t) on a storm basis and for the entire study watershed,
$Q$ is volume of runoff (in m$^3$),
$q_p$ is peak flow rate (in m$^3$ s$^{-1}$)
$K$ – the soil erodibility (in t ha h ha$^{-1}$ MJ$^{-1}$ mm$^{-1}$),
$L$ – slope length,
$S$ – slope steepness,
$C$ – crop management,
$P$ – soil erosion factor,
$a$ and $b$ are location coefficients.

For the areas where the equation was developed, $a$ and $b$ were 11.8 and 0.56, respectively, for metric system units. The optimization technique suggested in (DeCoursey and Snyder, 1969) was used for the development of the prediction equation and designating $a$ and $b$. A disagreement with the principle of dimensional analysis of the MUSLE has been explained in (Cardei, 2010). The MUSLE has been an attempt to estimate stream sediment yield for individual storms by replacing the rainfall factor with a runoff factor (Sadeghi et al., 2013). The MUSLE model produce reasonable estimates when it is applied under appropriate conditions as shows a significant difference with measured sediment yield in many watersheds. Applying of the model need calibration else in the result has huge errors.

But despite good results in some areas, for correctly applying of the MUSLE model is strictly recommended very careful review of the correct values and exact variables used.

**4.1.4 CLSE (Chinese Soil Loss Equation)**

The CLSE is empirical model was develop in (Liu et al., 2002) of USLE for China as estimating average annual soil loss by water on hillslope for cropland. The model predict interrill erosion from farmland under different soil conservation practices use an empirical multiplicative equation with six factors obtained from experiment stations covering most regions of China and modified to the scale of Chinese unit plot defined:

$$A = R * K * L * S * B * E * T$$

where,
- A – Annual average soil loss (t ha$^{-1}$),
- R – Rainfall erosivity (MJ mm ha$^{-1}$ h$^{-1}$ yr$^{-1}$),
- K – Soil erosibility (t ha h MJ$^{-1}$ mm$^{-1}$),
- S – Slope steepness dimensionless,
- L – Slope length factors dimensionless,
- B – Biological-control dimensionless factors,
- E – Engineering-control dimensionless factors,
- T – Tillage practices dimensionnes factors.

Chinese soil loss equation is to predict annual average soil loss from slope cropland under different soil conservation practices. In (Liu et al., 2002) the USLE model use 30 year data was adapted in substituting the C and the P factor with three factors considering the biology, engineering and tillage practices

**4.1.5 RUSLE** (Revised Universal Soil Loss Equation) **Model**

RUSLE is a straightforward and empirically based model that has the ability to predict long term average annual rate of soil erosion on slopes using data on rainfall pattern, soil type, topography, crop system and management practices. RUSLE is an empirical model, designed for use at runoff plot or single hillslope scales. However, erosion rates of ungauged catchments can also be predicted using RUSLE by using knowledge of the catchment characteristics and local hydroclimatic conditions.

In (Renard et al., 1997) is shown the Revised Universal Soil Loss Equation (RUSLE) model as an update version of USLE model. In the RUSLE model, the potential soil erosion risk consists of only the multiplication of **three natural factors** to indicate the area under high vulnerability:

- *Rainfall erosivity* (Nearing et al., 2017),
- *Soil erodibility*,
- *Slope length and Slope steepness.*

RUSLE can be expressed as:

$$A = R * K * L * S * C * P,$$

where,

    A = average annual soil loss per unit area (t ha$^{-1}$ yr$^{-1}$),
    R = rainfall-runoff erosivity factor (MJ mm ha$^{-1}$ h$^{-1}$ yr$^{-1}$),
    K = soil erodibility factor (t ha h MJ$^{-1}$ mm$^{-1}$),
    L = slope length factor,
    S = slope steepness factor,
    C = cover and management factor, and
    P = support and conservation practices factor (Panditharathne et al. 2019).

The RUSLE model is well studied and it has been widely applied at different scales to estimate soil erosion loss, and to plan erosion control for different land cover categories such as *croplands, rangelands, and disturbed forestlands.*

In comparison with the other methods such as **CORINE**, **NNIPHE**, **GLASOD** and Hot Spot approaches, as well as often uses data from Geographic Information System (GIS) the RUSLE model is considered as a leading model as gives the most detailed information on the soil erosion risks (Grimm et al., 2002; Claessens et al., 2008). Important disadvantage is that the available data for finding of some of the RUSLE parameters can be a limitation for obtaining of a maximal accuracy. Moreover, using of a model method often implies uncertainties in the calculation of each factor (Van der Knijff et al., 2000).

In (Fu et al., 2005) the RUSLE model has been applied at large watershed scale combined with GIS to assess soil loss using local data in the Yanhe. Because of the limitations of the RUSLE and spatial heterogeneity, more work should be done on the RUSLE R-factor and C-factor. Obtained is that the middle and southeast parts of the Yanhe watershed has more erosion than the northwest part. The main reason for soil loss is the close relationship with land use and rainfall–runoff erosivity. But, because of the limitation of RUSLE, spatial heterogeneity in the watershed and use of empirical data, there are uncertainties in the predicated value.

In (Kim, 2006) is analyzed the spatial distribution of' *soil erosion* in the Imha watershed, to determine the Sediment Delivery Ratio (SDR) in the Imha watershed and determine the Trap Efficiency (TE) at the Imha reservoir. To estimate the gross soil erosion as well as analyze the mean annual erosion and the soil losses caused by typhoon "Maemi" in the Imha watershed is used the RUSLE model combined with GIS techniques. The model is used to evaluate the spatial distribution of soil loss rates under different land uses. The RUSLE model and GIS techniques in (Bhat et al. 2017) to determine a quantitative assessment of average annual soil loss for Micro-watershed in J&K, India is considered. The results in the micro-watershed shows that the minimum rate of soil erosion is in the land areas with natural forest cover in the head water regions, higher rate of soil erosion - in areas with human intervention. Therefore this model helps to increase the prediction capability and accuracy of remote sensing and GIS based analysis. In (Winning and Hann, 2014; Lu et al., 2004) is applied the RUSLE, remote sensing, and GIS to the mapping of soil erosion risk in Brazilian Amazonia as the soil erodibility factor (K), and a digital elevation model image was used to generate the topographic factor (LS). In this research is that remote sensing by Landsat 7 satellite and GIS estimation and its spatial distribution feasible with reasonable costs and better accuracy in larger areas. Additional is shown the relationships between land use and land cover (LULC), and soil erosion risks which are useful for managing and planning land use that will avoid land degradation. As show as results a remote sensing and GIS provide useful tools for evaluating and mapping soil erosion risk in Amazonia.

**4.1.6 RUSLE 2** (Revised Universal Soil Loss Equation, version 2.0)

With the development and application of GIS technologies and expansion of computational power and refinement of mathematical expression, estimating erosion transitioned from an empirical to a process-based model. Although RUSLE2 is empirical model by nature, the model consists of the newest development of mathematical representations of tillage conservation and sub-factors of crop systems. RUSLE2 improves predictive ability by identifying extreme loss event using daily inputs. RUSLE2 is the replacement for RUSLE used commonly today in the United States by government agencies for conservation planning purposes where additional has been introduced the new concept of "*erosivity density*" for calculating and mapping rainfall erosivity in the United State (USDA-RULSE2, 2013). In (Nearing et al., 2017) is calculated er*osivity density* as an amount of erosivity per unit of rainfall depth, generally calculated on a monthly basis at a given location. The units of energy density are energy per unit time per unit area (MJ ha$^{-1}$ min$^{-1}$).These values are multiplied by average unit depth of precipitation (mm) for the period of interest (e.g., monthly) to give the average erosivity for the time period. Calculation of the erosivity density approach to rainfall erosivity provides better estimates of erosivity, smoother mapping across regions, better performance in mountainous areas, the capability to use shorter 15 year rainfall records, and the ability to utilize daily data in conjunction with less common 15-min rainfall data.

A major advancement in RUSLE was the use of sub factor relationships to compute C factor values from basic features of cover-management systems. RUSLE2 was developed primarily to guide conservation planning, inventory erosion rates and estimate sediment delivery (RUSLE2a). The equation is as USLE except it computes soil for a given day rather than an annual soil loss. The RUSLE2 contains several major enhancements including (RUSLE2b):

- Improved tools to develop and use annual and perennial vegetation descriptions,
- Tools to estimate runoff and develop a representative runoff event sequence,
- Tools to calculate and display tillage erosion on profiles with changing slope steepness,
- Tools to make it easier to apply and remove permeable barrier systems commonly used on construction sites.

The K-value has been made to fluctuate during the year, rather than remaining constant. The soil erodibility calculation method was also changed in order to express the soil structure sub factor. The difference between the USLE, RUSLE, and RUSLE2 results in the three methods giving different erosion estimates even when each method gives the same average annual values for each USLE factor. This difference results in as much as a 20% difference in average annual erosion values between RUSLE2 and the USLE and RUSLE. RUSLE2 displays a variety of erosion values that can be used in conservation and erosion control planning. Also, RUSLE2 can be applied in the traditional USLE way by assuming a uniform slope and that deposition ends slope length. The erosion values computed by RUSLE2 can be compared with soil loss tolerance values or other erosion control criteria just as USLE soil loss values were used.

In the equation of the factor K (*soil erodibility factor*) is used the *modified soil structure subfactor* $K_S$ given by (RUSLE2a):

$$K_S = 3.25(2 - S_t)$$

and in (Ismail and Ravichandran, 2008)

$$K_S = 3.25(S_t - 2)$$

where $S_t$ – soil structure class. A new subfactor has been added, and the deposition equations have been extended to consider sediment characteristics and how deposition changes these characteristics. In the RUSLE2 to estimate K-factor values for mixed soils is used modified soil erodibility nomograph.

RUSLE2 method works best where rainfall occurs regularly, rainfall is the dominant precipitation, and average annual rainfall exceeds 20 inches. Also can be used to estimate erosion in the special winter condition represented by the Northwest Wheat and Range Region, but in cases of snowmelt it does not explicitly estimate erosion.

A modified version of RUSLE model in (Panagos et al., 2015) with 100m resolution is constructed notation as RUSLE2015. It considers a negative impact on ecosystem services, crop production, drinking water, and carbon stocks. The impact of the Good Agricultural and Environmental Condition (GAEC) requirements of the Common Agricultural Policy (CAP) and the EU's guidelines for soil protection can be grouped under land management (reduced/no till, plant residues, cover crops) and support practices (contour farming, maintenance of stone walls and grass margins). RUSLE2015 was found to be the most suitable modelling approach for estimating soil loss at the European scale (in terms of validation, usability, replicability, transparency, and parameterisation) which means that can be very useful tool for simulating the effects of land use changes and land management practices on the rates of soil loss due to water erosion.

As is well–known RUSLE method is a computerized version of USLE with revised estimations of its equation factors which can be implemented of graphical processing units (GPU). Such an example is realized in (Sten et al., 2016) where two new parallel algorithms for flow accumulation calculations are presented. Need to note, flow calculations are a key quantity in many surface hydrological simulations as well as in the estimation of the LS factor (RUSLE method). The final implementation show that the calculation of RUSLE for an area of 12km × 24km (72 million cells, one UTM-25 mapsheet) in less than 1 sec. when using binary datasets and stored topological sorts. Therefore, the GPU implementation can be used in future as a system for erosion map production on large areas even whole countries.

**4.1.5 USPED** (Unit Stream Power Erosion Deposition) **Model**

The USPED is empirical model combines the USLE parameters and upslope contributing area to estimate sediment flow and then erosion and deposition rates are computed as change in sediment flow in the direction of steepest slope. As shown as in (Mitasova, 1996; Mitasova, 1999; Mitasova et al., 1996;) the model can to predict the spatial distribution of erosion, as well as deposition rates for a steady-state overland flow with uniform rainfall excess conditions. The model was applied to the complex topography of the catchment in order to obtain quantitative information on the processes of soil detachment and sediment deposition. Model spatial pattern estimates are similar to those of the RUSLE-3D model, showing consistently that erosion prediction is controlled by relief and land cover. The model predict satisfactorily the occurrence of erosion in areas where currently acute processes of erosion are taking place. However, the model fails to predict high erosion rates for areas where all arable land has already been long eroded.

**4.1.8 Zheng's Model**

Zheng's soil erosion model is described in (Zheng et al., 2008). It is empirical model develop a proportional function for event sediment yield prediction through analyzing the field observations at 12 small catchments over the Loess Plateau as well as updated through deriving the regression coefficient from the power function. The model can be effectively used as a management tool the vast majority of the annual sediment yield. In order to predict event sediment yield is proposed a proportional function. The performance of the model is good for high-magnitude events, especially extreme events.

**4.1.9 SYI (Sediment Yield Index) Model**

The SYI (Bali and Karale, 1977) is empirical model defined as the Yield per unit area and its value for hydrologic unit is obtained by taking the weighted arithmetic mean over the entire area of the hydrologic unit by using suitable empirical equation. It considering sedimentation as product of erosivity, erodibility and areal extent was conceptualized in All India Soil and Land Use Survey (AISLUS) as early as 1969 and has been in operational use since then to meet the requirements of prioritization of smaller hydrologic units. The erosivity determinants are the climatic factors and soil and land attributes that have direct or reciprocal bearing on the unit of the detached soil material. The SYI model is developed by AISLUS in order to identify critical hydrological units over a large basin, was tested in a drainage basin of the Western Ghats mountainous zone which receives heavy rainfall. It is a well known means of providing criteria for priority delineation in river valley projects and flood prone rivers (AISLUS, 1991).The SYI method is highly useful for prioritization of micro-watersheds according to erosion impact. The erosivity is simulated with the sediment yield weightage value which is based on assessment of the composite effect of assemblage of erosivity determinants:

$$\text{SYI} = \sum_{i=1}^{n} \frac{A_i \times W_i \times DL}{A_W} \times 100, \text{ for } i = 1,2,3,\cdots,n,$$

where,

$A_i$ – area of the erosion intensity $i^{th}$ unit (EIMU),

$W_i$ – weightage value of the erosion intensity $i^{th}$ mapping unit,

$n$ – no. of mapping units,

DL – delivery ratio,

$A_W$ – total area of sub-watershed.

The range in SYI values defined by the specific characteristics of the individual basin is divided of 6 classes – very high, high, medium, low, and very low.

In (Agegnehu et al., 2020) the two different empirical (RUSLE and SYI) models are compared and assessed the impact of rainfall-induced soil erosion and prioritize the sub-watersheds. Results show that two models of sub-watersheds have similar average soil loss at higher elevations. In this research are observed a strong antagonistic relationship between the elevation/slope zones and the mean sediment loss in both models. In (Gajbhiye et al., 2015) is shown a relationship between Soil Conservation Service Curve Number (SCS-CN) and SYI for the Narmada watersheds (Madhya Pradesh). The runoff curve number (CN) included as input for model development is in the interval CN = [0–100], but practically is used in CN = [40 – 98]. For the study area the SYI model predict the need to adopt the suitable soil conservation measure in the study watershed for minimizing the soil erosion. The simplified AISLUS model provides available catchment parameters as shows a good match with SYI model. In (Naqvi et al., 2015) the SYI method was used to calculate soil loss in micro-watershed. The values and thematic layers were integrated as per the model as well as calculate of minimum and maximum sediment yield values and classified into four priority zones according to their composite scores. Due to the important assessments if the model is need immediate attention for the study area and its conservation.

**4.1.10 G2 Model**

G2 is a soil erosion model for developing monthly erosion maps at regional scale was introduced by a team from Aristotle University of Thessaloniki and DG JRC within GEOLAND 2 project is included in JRC (G2, 2020). The G2 model proposes innovative techniques for the estimation of vegetation and protection factors. It is a complete, an empirical model for soil erosion rates and has evolved with time into *a quantitative model mapping* soil loss and sediment yield on month-time intervals, designed to run in a GIS environment. A detailed description in (Karydas and Panagos, 2018) is presented where the model adopts fundamental equations from RUSLE and the Erosion Potential Method (EPM). It aimed at developing an equation to estimate the water erosion and to put up a service for regional soil erosion monitoring across Europe (Karydas et al. 2012; Panagos et al., 2014; Panagos et al., 2015). The G2 model clearly shows the critical seasons, hotspots and land uses which are more susceptible to erosion as the model is described by the following formula (Panagos et al., 2014):

$$E = \frac{R}{V} \times S \times \frac{T}{l}$$

The *Dynamic factors*:

- E – The predicted soil amount removed from an area during a specific time period (t/ha$^{-1}$),

- R – The rainfall-runoff erosivity factor (modified from USLE by G2), which quantifies the impact of raindrop and runoff energy,
- V – The vegetation retention factor (dimensionless, and analogous to the USLE's *C*-factor), which represents the effects of all interrelated cover and management variables (developed by G2).

*The Static factors:*
- S – The soil erodibility factor (modified from USLE by JRC, 2000-5), identical to the USLE's *K*-factor, which reflects the ease of soil detachment by raindrop splash or surface flow;
- T – The topography factor (dimensionless and analogous to the LS-factor of the USLE), which expresses the effect of slope length and slope gradient (USLE modifications, 1996);
- I – Interception of slope length (developed by G2).

The formula was uptake by the community and was adopted for the whole territory of Greece. Develop and revise of the model as an effective decision-making tool is considered in (Karydas and Panagos, 2018) for study of areas in the South East Europe and Cyprus as in most cases (with exception of two cases – Crete and Cyprus), adequate field data were not available for a comprehensive accuracy. The month-time step assessments improve understanding of erosion processes, especially in relation to land uses and climate change. The results from use of G2 models show realistic features of flow conditions and consistent flow patterns. In (Halecki et al., 2018) the G2 model has been verified for the first time under Polish conditions. It was used to investigate soil erosion assessment for the monthly data in the Outer Western Carpathians, in the southern region of Małopolska Province, Poland. The result shows that the model can assess soil loss in cropland and forest-dominated land as well as in a steep-sloped agricultural basin with a variable hydrological regime. The assessing total soil eroded mass over monthly intervals enable determination the influence of plant growing stages on the V parameter. Therefore, due to use of various plant stages the model is important tool for soil and water conservation. In (Karydas et al., 2020) the model is used for the study of the Candelaro river basin in Apulia region (Italy) with Sentinel2 image scenes for the first time for erosion assessments. Sentinel-2 is a ready-to-use, image product of high quality, freely available by the European Space Agency. The results show that G2 model as a rapid, robust, and flexible mapping tool as shows an appropriate solution for erosion risk assessments in the whole Apulia region. Needed to note by using the G2 model are achieved to reveal the specific contribution of every land cover to soil loss as well as the seasonal changes of rain intensity and vegetation cover.

## 4.2 Conceptual models

A catchment is represented as a series of internal storages without including the specific details of process interactions. Parameters of conceptual models have limited physical interpretability which leads to its intermediate role between empirical and physically models (Sorooshian, 1991). Conceptual models can indicate the effects of land-use changes without requiring large amounts of spatially and temporally distributed input data.

**4.2.1 Thornes Model**

It is a conceptual erosion model (Ali and De Boer, 2010) that contains a hydrological component based on a runoff storage type analogy, a sediment transport component, and a vegetation cover component. It is based on square grid cells starts by an exponential frequency distribution within a specified time period. It is a geomorphic model that combines runoff rate, soil erodibility, the effects of topography, and vegetation protection in a simple physical equation. The Thornes erosion equation for each cell reads (Ali and De Boer, 2010):

$$E_i = \frac{kRO_i^2 s^{1.67}}{e^{0.07c_c}}$$

where,

$E_i$ is the erosion rate (mm month$^{-1}$),
$k$ – the soil erodibility coefficient,
$s$ – the slope (mm$^{-1}$),
$c_i$ – the fraction of vegetation cover (%),

$RO_i = \dfrac{P_i}{e^{\frac{rc-S_i}{P_i/D_i}}}$ - the surface runoff (mm),

$P_i$ – the total precipitation (mm),
$rc$ – the potential water storage capacity (mm),
$S_i$ is the total initial soil moisture (mm),
$D_i$ is the number of precipitation days, and $i = 1:12$ is the time period.

In (Zhang et al., 2002) the model is used for predicting global erosion rates which show that the model structure is suitable for predicting potential erosion rates at daily, monthly, and annual time scales. In (Thornes, 1985; Thornes, 1990) the model analyzes the competition between vegetation growth and soil erosion is constructed by combining sediment transport and vegetation protection. It contains a hydrological component based on a storage type analogy, a sediment transport component and a vegetation growth component.

**4.2.2 FLEX–Topo**

In (Savenije, 2010) a modelling approach rather than a new conceptual model structure is proposed for conceptual modelling approach enables us to design different model structures which are based on landscape units including plateau, hillslope and wetland with separate conceptualization of water balance, and constitutive equations. It is FLEX-Topo structure dramatically reducing the need for calibration by use of hydrological landscape analysis. The model uses a new topographical indicator Height Above the Nearest Drainage (HAND) as a topographic index to distinguish between landscape-related runoff processes which can be different for every climate, ecosystem, land-use system and morphological setting. The author opinion share the fact that the elements in FLEX-Topo structure are not connected which may be one of the key raisons the model structure to be conceptually better. Fig. 1 shows the three sub-systems – wetland, hillslope, and plateau as well as they limits where details they conceptual model structure that should reflect the structure of these sub-systems in the real world.

It is simple conceptual approach to hydrological modelling, where the topography linked to geology, geomorphology, soil, land use, ecosystems, climate, is uses for classification. The main advantage of an approach is its maximum simplicity, which including the observable landscape characteristics.

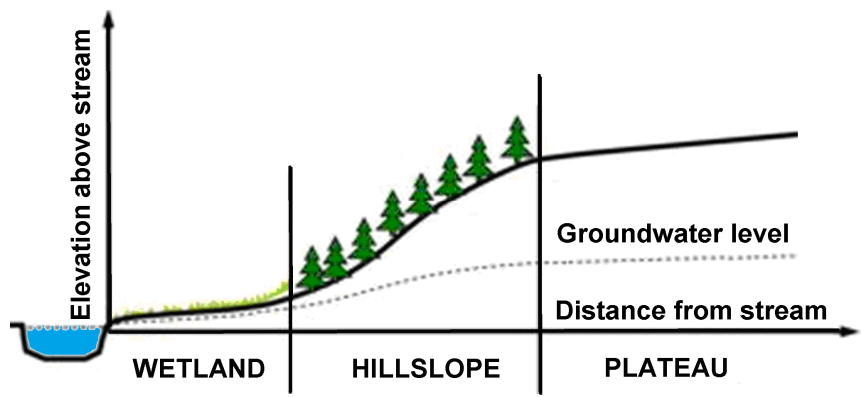

Fig.1 FLEX-Topo model concept: The thee sub-systems – wetland, hillslope, and plateau and its limits (Savenije, 2010).

In (Gao et al., 2014) FLEX-based approach is used for tested of four models – $FLEX^L$, $FLEX^D$, $FLEX^{T0}$, and $FLEX^T$ where the study area is a tailor-made hydrological model for a cold, large river basin in north-west China. The model $FLEX^L$ is lumped, the $FLEX^{T0}$ and $FLEX^D$ models are semi-distributed, and $FLEX^T$ model is with the structure and parameterization as $FLEX^{T0}$. The results from the four models shows that $FLEX^{T0}$ and $FLEX^T$ models are better spatially transferable than others, as well as a better transferability was performs by using $FLEX^T$. For represent the heterogeneity of hydrological functions more appropriate are $FLEX^{T0}$ and $FLEX^T$ models. It is means that they have a more realistic model structure and parameterization.

**4.2.3 MMF** (Morgan-Morgan-Finney) **Model**

The MMF model developed in (Morgan et al., 1984) is a *process-based (physically based model)*, spatially-distributed consisting from following two phases:
- ➢ *A water phase* – runoff volume is estimated as an exponential function of rainfall, with a consideration of vegetation interception, topography, soil water storage and routing
- ➢ *A sediment phase* – the model estimates rainfall splash via rainfall energy and interception and transport capacity of runoff based on runoff volume, slope gradient and crop management.

The basis for development of physically-based (or *process-based*) soil loss model is the mass balance differential equation which in 1D a hill slope profile is:

$$\frac{d(cq)}{dx} + \frac{d(ch)}{dt} + S = 0$$

where,

$c$ – Sediment concentration (kg m$^{-3}$),
$q$ – Unit discharge of runoff (m$^2$ s$^{-1}$),
$h$ – Depth of flow (m),
$x$ – Distance in the direction of flow (m),
$t$ – Time (s),
$S$ – Source/sink term for sediment generation [kg m$^{-2}$ s$^{-1}$].

Annual sediment yield is determined as the lesser of the amount of soil-particle detachment and transport capacity. The model predicts annual soil loss from hillslopes. The obtained results in (Morgan et al., 1984) are with respect to examine the effects of shifting cultivation on soil erosion in a tropical rainfall forest from Malaysia show that, except for very low and very high rates of erosion, realistic predictions over a wide range of conditions are founds. The model there is some improvements included in its revised versions (De Jong, 1994; Morgan, 2001; Morgan and Duzant, 2008).

As know well Earth observation and GIS technology enables us to extract information from satellite images and from digital elevation models (DEM) and to process vast amounts of data. This gives a possibility to be construction in (De Jong et al., 1999) modified MMF method using of a cumulative sediment transport capacity algorithm, remote sensing imagery and DEM. The model estimates annual soil loss by evaluating both rainfall soil detachment and sediment transport over the soil surface.

The GIS and MMF model is applied in (Tesfahunegn et al., 2014) for estimating soil erosion in the Mai-Negus catchment, northern Ethiopia where the input data include climate, topography, land use, and soil data. The MMF modelling processes erosion in two phases (Morgan et al., 1984):

- *The water phase* – The water phase mainly comprises of prediction of soil detachment by rain splash as requires data for intensity of rainfall, number of rainy days, and average annual rainfall.
- *Sediment phase* – soil erosion to result from the detachment of soil particles from the soil mass by raindrop impact and the transport of those particles by overland flow.

The model estimated soil loss at catchment level was compared with the survey based measured sediment yield from the reservoir located at the outlet of the catchment. Since the results from MMF model shows a lower rate of erosion for the soil transport capacity of over land flow it follow that erosion is transport limited. Also is important to note integrate of GIS is a useful tool to manage spatially distributed hydrophysical data while assessing the spatial distribution of erosion.

*4.3 Physically–based models*

Physical models are based on fundamental physical equations and their solutions describe sediment and stream flows in a catchment. The models represent controlling erosion and sediment yield as well as physical characteristics, topography, geology, land use, climate, plant growth and river flow characteristics.

**Advantages** of physically-based models in comparison with empirical/conceptual models
- more accurate extrapolation to different land use;
- more correct representation of erosion/deposition processes;
- application to more complex conditions including spatially varying soil properties and surface characteristics;
- More accurate estimation of erosion/deposition and sediment yield on a single storm event basis.

**Disadvantages** are such as large data requirement, lack of user-friendliness, unclear guidelines for conditions of their applicability, improper measure of reliability and lack of expression of their limitations.

### 4.3.1 SWAT (Soil and Water Assessment Tool) Model

It is a river basin, or watershed, process – based scale model developed by the United States Department of Agriculture–Agricultural Research Service (USDA–ARS) (Arnold et al., 1998; Srinivasan et al., 2004) to predict the impact of land management practices on water, sediment and agricultural chemical yields in large complex watersheds with varying soils, land use and management conditions over long periods of time. It requires specific information about weather, soil properties, topography, vegetation, and land management practices. It is a physically-based model to estimate runoff, nutrient losses, chemical and sediment transport within the watershed scale for daily time step. SWAT is a continuous time, physically based hydrological model based of the *soil water balance equation* (Landsberg and Sands, 2011) calculated as the month is divided into $d_R$ equal periods with an amount $\frac{R}{d_R}$ rain falling as a single event at the beginning of each period, where $R$ is the total monthly rainfall and $d_R$ the number of rainy days, and the water balance is performed separately over each period. The hydrologic routines within SWAT model account for vadose zone processes (i.e., infiltration, evaporation, plant uptake, lateral flows, and percolation), and ground water flows. Sediment transport is a function of deposition and degradation, which are determined through comparing the sediment concentration and maximum sediment concentration. But the method does not account for rejected recharge. In (Neitsch et al., 2011) is shown that the method is computationally efficient and enables users to study long-term impacts as simulation of very large basins, a variety of management strategies can be performed without excessive investment of time or money. The model is not designed to simulate detailed, single-event flood routing.

Together with satellite remote sensing and GIS this model in (Mosbahi et al., 2013) show that can be very useful tools to estimate surface runoff, soil erosion, predict surface runoff generation patterns and soil erosion hazard. Also, the method is effective for identifying and prioritizing vulnerable sub-catchments. The SWAT model includes presently a variety of parameters for which there is no information available. In order to focus on the most significant ones, a systematic sensitivity analysis is needed. In (Karki et al., 2020) is considered in details the methods used for field-scale SWAT modeling and discusses the limitations and advantages.

**4.3.2 QSWAT** (Quantum Soil and Water Assessment Tool) **Model**

The QSWAT is process-based method an implementation of the SWAT in Quantum GIS (QGIS). It is improved version of SWAT model developed in (Dile et al., 2016) as open-source software set up with Python uses various functionalities of QGIS and is installed as its plugin. QGIS is a free and open source desktop GIS application that provides data viewing, editing and analysis capabilities. The model was input with various watershed parameters as spatial variation of topographic features along with weather parameters. QSWAT has the capability to visualize results. It can help visualize static data, and it can animate the results at model simulated time steps. It demonstrates successfully with Gumera watershed located in Lake Tana basin, tropical highland region of Ethiopia. QSWAT model has improved the processing capabilities of the SWAT and lesser processing time on Digital Elevation Models (DEMs), and better statistical and dynamical representation of outputs.

**4.3.3 DYRIM** (Digital Yellow River Model)

DYRIM is created in (Wang et al., 2007) to be a framework to simulate the dynamic erosion, transport, and sedimentation processes at a range of spatial and temporal scales for the Yellow River. This is a process-based, physically based spatially-distributed and continuous model for land management and to predict soil erosion and sediment transport for Loess Plateau catchments. The sediment yield equation is derived by simple assumptions where sediment discharge would increase nonlinearly with the slope length, leading to unrealistic results as the spatial scale increases. It was used successfully to simulate the huge soil loss (Guo et al., 2015). Some improvements of the model are shown in (Guo et al., 2015) as well as the parallelization to improve its computational efficiency and parallel algorithms have been implemented on high-performance clusters in (Li et al., 2011; Wang et al., 2013; Wu et al., 2013).

**4.3.4 EPIC** (Erosion Productivity Impact Calculator) **Model**

The EPIC model (Sharply and Williams, 1990; Williams et al. 1984; Williams, 1990) was originally developed to determine the relationship between soil erosion and soil productivity throughout the U.S. The EPIC is a physically based model which components include weather simulation, hydrology, erosion-sedimentation, nutrient cycling, plant growth, tillage, soil temperature, economics, and plant environment control. The EPIC model is operational and has produced reasonable results under a variety of climatic conditions, soil characteristics, and management practices. It has also demonstrated sensitivity to erosion in terms of reduced crop production. It was used for that purpose as part of the 1985 RCA (1977 Soil and Water Resources Conservation Act of USA) analysis. To simulate rainfall/runoff erosion, EPIC contains six equations – the USLE, the Onstad-Foster modification of the USLE, the MUSLE, two recently developed variations of MUSLE, and a MUSLE structure that accepts input coefficients. The approach in EPIC estimates potential wind erosion for a smooth bare soil by integrating the erosion equation through a day using the wind speed distribution.

**4.3.5 LISEM** (Limburg Soil Erosion) **Model**

The Limburg Soil Erosion Model (LISEM) (De Roo et al., 1996; De Roo and Jetten, 1999) is a process-based and a distributed physically based hydrological and soil erosion model developed for planning and conservation purposes. LISEM incorporates a number of different processes, including rainfall interception, surface storage in micro-depression, infiltration, vertical water movement through the soil, overland flow, channel flow, detachment by overland flow and transport capacity of flow. LISEM does not simulate concentrated erosion in rills and gullies; rather it simulates flow detachment only in the ponded area. This can be seen as intermediate between sheet and rill erosion. Processes describing sediment detachment by rainfall, through fall and overland flow are included, in addition to the transport capacity of the flow.

**4.3.6 ANSWERS** (Aerial Non-Point Source Watershed Environment Response Simulation)

The ANSWERS model developed in (Beasley et al., 1980) presented one of the first operational, fully spatially distributed, catchment erosion and sediment yield models. It is a physically based model that simulates excess rainfall, the orientation of laminar and rill flow, subsurface drainage, and the removal and transport of sediment at the watershed scale, applicable to basins with an area of up to 100km$^2$. The model is limited in the size of watershed it can deal with. The ANSWERS model consists of a water erosion and sediment transport model described in (Beasley et al., 1980) as:

- *Hydrologic*

The elemental size is defined such that the pertinent hydrologic and erosional variables can be assumed to be uniform within the watershed element where conditions may vary greatly from one element to the next. A typical cross-sectional view of a small area or element within a watershed is depicted in Fig.2.

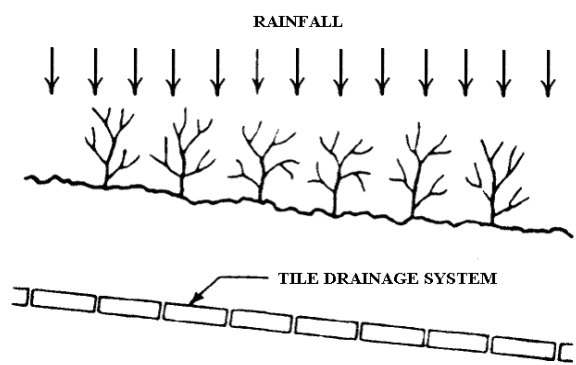

Fig.2 Profile of a watershed element (Beasley et al., 1980)

- *Erosion*

Soil detachment, transport, and deposition are closely related to the concurrent hydrologic processes occurring in a watershed. Detachment and transport can be accomplished by either raindrop impact or overland flow. However, the small amount of sediment transported from a field by raindrop impact was neglected. The processes described in the erosion model are shown in Fig. 3.

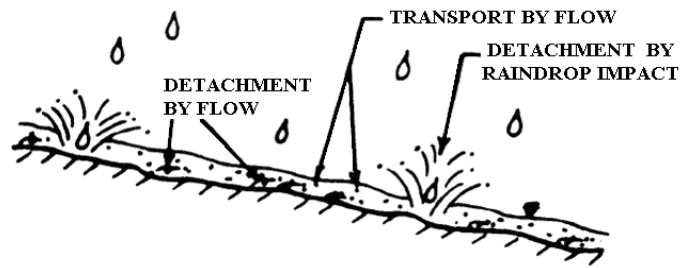

Fig.3 Sediment detachment and transport (Beasley et al., 1980)

The ANSWERS program is a comprehensive model intended to be used in quantitatively evaluating nonpoint source pollution in an ungaged watershed and in determining the relative effectiveness of alternative corrective plans. The distributed analysis also provides a characterization of the hydrologic response as well as erosion and deposition occurring throughout the watershed during a storm event. The disadvantage of the model is that it requirements for large amount of input data and analyze of results.

**4.3.7 CREAMS** (Chemicals, Runoff and Erosion from Agricultural Management Systems)

CREAMS model in (Knisel, 1985) is a physically based and a field-scale (less than 5 ha in size) model uses various aspects of the USLE equation, that predicts runoff, erosion and chemical transport from agricultural areas operating as well as in single storm events as in a long-term average mode. The model includes four components: hydrology, erosion, plant nutrients, and pesticides. The erosion component of CREAMS simulates detachment of soil particles and subsequent transport of these particles. Hydrologic inputs include rainfall depth, rainfall erosivity, runoff volume, and peak runoff rate. Soil-water content is calculated from a water balance that includes evapotranspiration, percolation, seepage below the root zone, and snowmelt.

**4.3.8 PESERA** (Pan-European Soil Erosion Risk Assessment) **Model**

The model has been developed from earlier models (Kirkby and Neale, 1987; Kirkby and Cox, 1995) based conceptual separation of precipitation into over-land runoff generation and infiltration, with a runoff threshold depending primarily on soil and vegetation properties. It is a physically based, spatially distributed, long-term coarse scale soil erosion risk model across Europe at a spatial resolution of 1000m in (Kirkby et al., 2008) and even 250m in (Berberoglu et al., 2020). It combines the impacts of soil, climate, vegetation and topography and soil erosion. It is a useful tool for soil and water conservation and informing ecological restoration. The model explicitly describes the processes of hydrology, vegetation growth, erosion and their interactions and therefore has a robust theoretical basis. It has been widely applied and validated across Europe and is currently used by planners and policy makers to determine funding decisions for soil erosion protection measures on farms. The model also assists in understanding the links between different factors causing erosion as well as scenario analysis for different land use and climate. The model considering 1D hydrological balance that partitions precipitation between evapotranspiration, overland flow, subsurface flow and groundwater recharge. It has been designed to produce state-of-the-art soil risk evaluation at a European scale where soil erosion is estimated as the amount of sediment carried to the base of a hillside land delivered to the channel network.

The model used with data for Turkey area in (Berberoglu et al., 2020) indicated that climate change increase soil erosion due to the changes in the precipitation regime and temperature rise. However, biomass increase as a result of temperature increase and precipitation in some regions will decrease soil erosion. Modelling soil erosion in the long term will enhance our understanding of the spatial variation of soil erosion to device soil conservation schemes.

In (Karamesouti et al., 2016) is shown predicting post-fire soil erosion loss where RUSLE model predicted higher soil losses than PESERA probably because only considers the sediment transport processes.

The model shows in (Tsara et al., 2005) a clear and explicit dependence on vegetative factors and to the specific rain fall regimes within Zakynthos, Greece, as is used for a regional diagnostic tool under a range of soil, topographic and climatic conditions for identifying the best land use type and vegetation cover to protect hilly areas from soil erosion. Comparison between the data obtained from the PESERA model and the measured values in the various soil erosion plots generally showed a satisfactory performance by the model.

In (Licciardello et al., 2009) is evaluate the sediment transport model to predicted monthly erosion rates were also calculated using observed values of runoff and vegetation cover instead of simulated values. By using of the multistep approach PESERA model shows promise to predict annual average spatial variability quite well. The results show the model allows prediction of the average annual spatial variability of runoff and erosion rates quite well over a range of land-use/land management systems in two contrasting climates. The advantage of the PESERA modelling framework is that it describes runoff and sediment transport characteristics and allows more realistic incorporation of effects of global change. Therefore this model can be used to predict runoff and erosion with reasonable accuracy based on limited input data.

**4.3.9 WEPS** (Wind Erosion Prediction System) **Model**

Soil erosion by wind is initiated when wind speed exceeds the saltation threshold velocity for a given field condition. After initiation, the duration and severity of an erosion event depends on the wind speed distribution and the evolution of the surface condition. Because WEPS is a continuous, daily, time-step model, it simulates not only the basic wind erosion processes, but also the processes that modify a soil's susceptibility to wind erosion as assumes a flat topographical field surface. By changing the field input parameters in for a specific field of interest can be compared various alternatives to control soil loss by wind.

The WEPS is a computer based model developed to provide an accurate, universal, and simple tool for simulating soil wind erosion. The strongest benefit of WEPS is its ability to provide producers a system to apply different "*what-if*" management scenarios to the land for developing and evaluating alternative wind erosion control practices. The WEPS is a process-based (USDARS, 1995; Wagner, 2013; Maurer and Gerke, 2011; Funk et al, 2004; Coen et al., 2004), daily time-step computer model that predicts soil erosion by simulation of the fundamental processes controlling wind erosion. WEPS represents new technology in wind erosion and is not merely an improvement in Wind Erosion Equation. It is designed for conservation planning and application providing the user with many elements of wind erosion including soil movement, estimated plant damage, and PM-10 emissions. WEPS replaces the predominately empirical Wind Erosion Equation (WEQ) (Woodruff and Siddoway, 1965) as the wind erosion prediction tool and can to predict average erosion along line-transects across the field. WEQ predicts average erosion along line-transects across the field, whereas WEPS treats the field as two-dimensional. The WEPS EROSION submodel simulates soil loss/deposition at grid points over the entire simulation region. WEPS model and calculates soil loss on a daily basis and analyze the output data to determine the times of the year when conservation treatments may be needed. WEPS model is coded in FORTRAN in (Carey et al., 1989) and last version 2012.8 was released on October 3, 2012 (Wang et al., 2010).

In (Hagen, 2007) is presented some update equations in the model which also enable direct simulation of field - scale fetch effects, and the durability of protective clods, crusts, and surface roughness. Estimating erosion effects on surface immobile elements as well as the buildup of mobile soil on downwind surfaces at the end of storms allow right and fast decision-making solutions.

**4.3.10 WEPP** (Water Erosion Prediction Project) **Model**

It is a process-based, spatially-distributed model. Compared with the USLE method (Flanagan et al., 2007) which was used extensively to predict long - term average annual soil loss, it was a mature technology was only applicable to detaching regions of a hillslope, and could not estimate sediment deposition or sediment delivery from fields to off - site channels or streams. Also USLE method no capabilities to estimate runoff, spatial locations of soil loss on a hillslope profile or within a small watershed, channel erosion, effects of impoundments, recurrence probabilities of erosion events, or watershed sediment yield. Developed to address all of these needs leads to the WEPP (Nearing et al., 1989) as serve empirically based erosion prediction technologies like USLE as well as software development including GUI and integration of WEPP with GIS software. It is a physically based model with distributed parameters that can be used in either a single event or continuous time scale and calculates erosion from rills and inter-rills, assuming that detachment and deposition rates in rills are a function of the transport capacity.

A more process based model of rainfall erosion than RUSLE2 is WEPP presented in (Tiwari et al., 2000). It is used to predict soil loss on an event basis using rainfall events that are generated stochastically over many years by a CLIGEN model (Nicks et al., 1995) but does not take account of gully erosion. Given that rainfall is not distributed evenly through time, it can be argued that the event soil loss predicted by WEPP is more realistic than predicted by RUSLE2. CLIGEN can be used as a weather generator for RUSLE2 to in order to predict long term erosion under conditions where rainfalls do not occur at regular intervals during the calendar year.

It is used for runoff and erosion modelling at a field or small catchment-scale (< 260 ha) (Laflen et al., 1991; Ascough et al., 1997) and in (Dun et al., 2009; Flanagan et al. 2012) has updates for different conditions and expansion of capabilities. Catchment-scale allows investigation of the connectivity between hillslopes and channels, and the seasonal and interannual role of channels on sediment accumulation and remobilization. Here by using of the sediment continuity equation are modeled detachment transport and deposition within channels.

In (Bulygina et al., 2007) is considered a new dynamic. WEPP (DWEPP) provides a new tool for assessing erosion rates and dynamics. It has physically based erosion mechanics descriptions and is sensitive to treatment differences on the experimental plots and has a well developed parameter database inherited from WEPP. The results shows that both WEPP and DWEPP gave a similar level of prediction accuracy for total event soil losses measured from both rainfall simulation and small watershed experiments.

**4.3.11 WATEM/SEDEM** (Water and Tillage Erosion Model/ Sediment Delivery Model)

It is a process-based spatially-distributed model (Van Rompaey et al., 2001) for the prediction of sediment delivery to river channels has three main components:
- Assessment of the mean annual soil erosion rate(2D RUSLE method);
- Assessment of a mean annual transport capacity as a function of the potential rill erosion;
- By using of the topography and transport capacity construction of an algorithm that routes the sediment along a continuous flow path towards the river system.

In later version of the model include derivation of the transport capacity to consider gully erosion (Verstraeten et al., 2007), and include of the slope gradient factor for steep slope erosion in (Zhao et al., 2015). Application of the model about the Chinese Loess Plateau in (Feng et al., 2010) show that although the modified version soundly estimated annual sediment yield for a small catchment, it is producing unsatisfactory/unstable predictions and negative Model Efficiency (ME) (Nash and Sutcliffe, 1970), Due to limitations in USLE/RUSLE models for quantifying soil erosion rates in data poor regions, erosion values of the WATEM/SEDEM model in (Delgado and Canters, 2012) are used for predicted and calculates how much sediment is produced onsite from water erosion for each catchment, soil erosion. Moreover, because of that the model incorporate the spatial pattern and composition of different land automatically is calculated the LS factor using the DEM of the catchment, the parcel map and the road network map as inputs. The obtained risk scores are catchment-specific where less than the mean catchment value indicate low relative soil erosion risk, while values higher than the mean indicate higher than average risk. The results obtained from model show that the areas to erosion are in zones where steep and longer slopes occur, without agroforestry or tree cover to impede run-off. The use of the model allowed us to predict the impacts agroforestry systems and other land uses have on the spatial patterns of soil erosion risk in three catchments of Claveria, the Philippines. The results suggest that upland areas in the tropics should promote the practice of agroforestry in areas prone to soil erosion.

When the soil erosion models are projected to consider impact of climate is needed, its choice must be forced by both precipitation and runoff. It is considered in (Eekhout and Vente, 2019) for Mediterranean catchments (south-east Spain) with models use different concepts: RUSLE forced by precipitation, MUSLE – forced by runoff, and MMF – forced by precipitation and runoff. The results are – RUSLE model show a decrease in soil loss due to a decrease in the annual precipitation sum, MMF – a projected decrease in detachment by raindrop impact is counteracted by a projected increase in detachment by runoff, and MUSLE – an increase in soil loss due to an increase in extreme precipitation, increased runoff. Therefore, we can use soil erosion models to best assess the impact of climate change on soil erosion. Some applications for soil erosion of the WATEM/SEDEM model in Europe (Borrelli et al., 2018).and estimated the possible phosphorus losses due to water erosion in Czech Republic (Krasa et al., 2019).

## 5. OUTLOOK ON ARTIFICIAL INTELLIGENCE BASED METHODS AND NEURAL NETWORK FOR SOIL EROSION RISK ASSESSMENT

In recent years, Artificial intelligence (AI) (Shahin, 2016) and machine learning (ML) (Mokhtarzad et al., 2017; Quej et al., 2017; Zhou et al., 2017) approaches, Deep Learning (DP) (Kamilaris and Prenafeta-Boldú, 2018), Support Vector Machine (SVM) (Schölkopf et al., 1999), Adaptive Neuro – Fuzzy Inference System (ANFIS) (Jang, 1993), and Artificial Neural Network (ANN) have been applied to modeling and analyze of the complex behavior of most geotechnical engineering applications as well as data processing from plot experiments for the purpose of soil erosion prediction. Machine learning based models can provide a helpful alternative to deal with the multivariate and complex nature of the phenomena in soil science and geoscience as well as achieve of better results with computational systems based on FF – NN (feed-forward neural networks). Since ANN methodology is based on empirical risk minimization models are trained by steepest gradient descent method with error backpropagation which means that the training phase of these models can be easily trapped in local optimal solutions as is shown in (Ngo et al., 2018).

As early as 90's of $20^{th}$ century to forecast the soil erosion is used Kohonen's self-organization neural network (SONN) model which is a two-layer network – each input node is connected to every output node via a variable connection weight (Cai, 1995). The results from prediction are totally in conformity with the actual results. Therefore NN theory can be very useful and fast tool for the soil erosion predictions.

### 5.1 ANN (Artificial Neural Networks)

An example of the available AI technique is shown in (Azizi and Gilandeh, 2020) where are used real world images as inputs of convolutional neural network (CNN) for classify the soil aggregates where is obtained good performance of the deep learning method as well as robustness of the presented deep model to noises and other unwanted signals. In (Achieng, 2019) is proposed machine learning algorithms (artificial neural networks (ANN) and deep neural networks (DNN)) for soil water retention curve (SWRC) for analyzing of flow and contaminant transport in the vadose zone. Results show that for loamy sand subjected to either monotonic wetting or drying, the DNN model outperformed ANN, while the RBF-based support vector regression (SVR) outperforms all the machine learning techniques. For a quantitative analysis of soil loss, ANN is applied in (Licznar and Nearing, 2003) where is found that ANN is generally better than the WEPP model. Later, in (Kim and Gilley, 2008) again by applying of ANN is estimated soil erosion where the results from a model show that the amount of soil eroded is positively correlated with rainfall and runoff.

Moreover, ANN in (Aslan et al., 2019) together with an erosivity index (EI) is used to predict potential soil erosion as a function of the precipitation regime and periods of aridity. The divided of 4 zones according to the risk of soil erosion and by using of different predictions are obtained erosion risk areas in Turkey for the years 2015–2040 based on an ANN model. It is very important because of, as shows as results, the low-risk are in northern and central Turkey areas while high risk is found in western and southern Turkey. It is shows the application of ANNs in geoscience can be as well as can be extended to climate change as very important further applicability.

Some machine learning algorithms is considered and compared in (Huang et al., 2020) for landslide susceptibility prediction (LSP) and the effect of the soil erosion models for Ningdu County, China. The results indicate that the soil erosion models have more accurate LSP than the single models with no soil erosion. The support vector machine (SVM) models (Bui et al., 2016), based on statistical learning theory, show higher prediction performance than the multilayer perceptron (MLP) model, using is a feed-forward neural network, and Logistic Regression (LR) (Tsangaratos and Ilia, 2016) model, using a multivariate statistics.

In (Albaradeyia et al. 2011) the study area the central highland mountainous of the Palestinian territories for the prediction of runoff and soil loss is considered. By compared results of WEPP method and ANN is shown that the soil erosion is highly dependent on both the rainfall depth and the rainfall event duration rather than on the rainfall intensity. Moreover results from the WEPP model for the soil loss and runoff disagree with the field data which means that this model is unsuitable by consider of such study areas. the ANN show a relatively unbiased estimation for both runoff and soil loss which means that it could be used as a management tool for predicting runoff and soil loss.

In (Abdollahzadeh et al. 2011) to predict soil erosion are compared ANN with Multi Linear Regression (MLR) for artificial rainfall data in a catchment located in Jakujo Rachidani in Tanakami area, in central Japan. One-hour recorded data from artificial rainfall test was used for current study and divided into two categories for training and testing: 70 % for training and 30% for testing phase. The ANN model was trained by outcome data of runoff and rainfall based on each minute data from artificial rainfall test. The obtained results show high ability of ANN to prediction of soil erosion as well as better matching between observed soil erosion and predicted data shows the high forecast capability of ANN model compared to MLR model.

In (de Arruda et al., 2016) is use ANN to a preliminary soil classification of an area not previously mapped using mapping units. The study area is the Barra Bonita municipality, state of São Paulo, Brazil where is needed obtaining of a digital soil map and environmental variables that express soil-landscape relationships. Considered environmental variables – elevation, slope, slope pro-file, slope plan and convergence index, are derived from a digital elevation model (DEM). Obtained results show an accuracy level of 82% between the reference points and the digital classification from ANN. The obtained soil map by the NN was in agreement with the map made for the reference area, and follows the same soil-landscape relationship and soil-parent material. This means that ANN can be apply to recognition of soil classes in areas where pedological surveys are not available.

The able on an ANNs to spatially predict in (Behrens et al. 2002) is used for digital soil map was predicted. A study area Palatinate Forest is located in southern Rhineland-Palatinate between Kaiserslautern and the German-French border. The integration of ANN with a GIS provides the opportunity to predict soil units. Appling of ANN there is two stages:

- The first stage – the network is trained where the output unit represents the target variable as the desired output;
- The second stage – the learned knowledge by weights is applied to prediction areas, for which the same input parameters (i.e., terrain attributes, land use, and geological units) are available, but no soil map has been surveyed.

Since predicts the soil units based on the learned weights. Validation returned provides remarkable results where a mean accuracy of ANN is over 92%.

In (Bagheri-Bodaghabadi et al., 2015) multilayer perceptron (MLP) artificial neural networks (ANNs) were developed with terrain attributes to map soil units using digital elevation model (DEM) attributes. The study area was located in the Borujen region of Chaharmahal-Va-Bakhtiari Province, Central Iran and has two dominant landscape units: hill-land and piedmont. The terrain attributes (slope (angle), aspect, curvature, maximum and minimum curvature, planform curvature, profile curvature, tangent curvature, wetness index, power index, sediment index, solar radiation, direct radiation, diffuse radiation and direct duration) were derived from a DEM. At the lowest level, smaller errors were observed with the World Reference Base (WRB) classification criteria than the Soil Taxonomy (ST) system, but more soil classes could be predicted when using ST. As expected, soil prediction using a higher level of classification presented a better overall level of accuracy.

The performance of different types of analytical models such as USLE, RUSLE can handle only five variables while in machine learning approaches there are many possibilities. In machine learning algorithms feature selection is an important method used in preliminary phase of model building. Before the model training phase, it is essential to judge the relevance of the selected factors used for landslide susceptibility mapping.

For seek the optimal method to map gully-prone areas there was compared the performance of three machine learning algorithms, Fisher's linear discriminant analysis (FLDA) (Bal and Örkcü, 2007) where all the samples are processed using linear projections, and a series of aligned scatter points, logistic model tree (LMT) (Landwehr et al., 2005) which is combination of logistic regression and the classic decision tree model and its results are compared to other machine learning models with low variance and bias, and Naïve Bayes tree (NBTree) as well as introduce three novel ensemble models by combining the base classifiers to the Random SubSpace (RS) which was initially proposed by (Ho, 1998) and to create ensemble models for classification and regression trees, meta-classifier – RS-FLDA, RS-LMT, and RS-NBTree.

The NBTree model is a hybrid from the most widely used approaches to solve classification problems- the Naive Bayes (NB) and decision tree DT models (Wang, 2019) as combines the advantages of the two models as possesses notable statistical significance.

The results indicate that hydrology-related factors contribute the most to gully formation and investments can be made to control overland water fluxes. This also is confirm by the RS-NBTree model which outperforms than the other five models, as indicated by the prediction accuracy, goodness-of-fit, and possess greater flexibility towards unknown data. The model is able to adapt and predict an unknown dataset.

In (Padarian et al., 2019) deep learning (DP) is applied to learning the convolutional neural network (CNN) model leads to a significantly improve of the prediction accuracy over conventional digital soil mapping models in Chile. DP is a machine learning type method which learning the representation of data through a series of processing layers. The model is characterized with some unique features – a 3D stack of images, data augmentation to reduce overfitting, and the simultaneous prediction of multiple outputs. Moreover, the DP model simultaneously predicts soil organic carbon at multiples depths with less prediction uncertainty and a reduced error with 30% compared with conventional techniques.

## 5.2 RF (Random Forest) Model

Random Forest (RF) (Breiman, 2001) is obtained by an ensemble of uncorrelated, simple regression trees (Breiman et al., 1984) as the classifier (Quiroz et al., 2018) is machine learning-based. The algorithm is a tree type classifier combines the bagging ensemble learning and random subspace. It is an effective decision tree ensemble used for large-scale and multivariate pattern recognition. The core function includes:

- Random selection of the sample features 'k' from a total sample 'm', where 'k' < 'm',
- Calculation of the node tree 'd' from the features 'k',
- Split '*d*' into daughter nodes tree 'dn',
- Repetition of the above steps untill 'I' the number of nodes tree is achieved.

It makes it a powerful tool for pattern recognition of large-scale and multivariate data as some decides about maximum nodes to its output class are considered in (Bonissone et al., 2010). In (Breiman, 2001) was extended as is introduced the concept of bagging and random feature selection and application including landslide modeling is shown in (Chen et al., 2018). In (Phinzi et al., 2020) assess soil erosion risk by two methods - RUSLE and Random Forest (RF) in the Umzintlava catchment river and its tributaries, the Eastern Cape (EC) province, South Africa is considered. In terms of commission error RUSLE method show overestimated soil erosion with about 16% (of non-erosion area classified as erosion) which means that a method still has limitations and weaknesses in certain regions. Some discrepancies between hotspots of two methods are observed. The results indicate that the integration of two methods successfully delineated the spatial patterns of soil erosion, providing useful information on erosion risk but respect to kappa coefficient RUSLE-based results shows very low levels of accuracy = 57%, while RF-based accuracy = 85%. Such low accuracies of the RUSLE model can be assigned of underestimated soil erosion in areas of gentle slopes within the catchment. The results achieve highest accuracies are obtained for RF-derived erosion. Random forest (RF) classifier with combination with internal metric of mean decrease accuracy (MDA) is used in (Šandera and Štych, 2020) to select the most suitable predictors to detect changes from permanent grassland to arable land. Biological predictors, such as leaf area index (LAI), fraction absorbed photo synthetically active radiation (FAPAR), normalized difference vegetation index (NDVI), are tested in the form of a time series from the Sentinel-2 satellite. The proposed change detection procedure achieved a very high accuracy. Consequence, results shows the RF classifier and its internal MDA metric are able to successfully detect relevant predictors for changes from permanent grassland to arable land.

**5.3 WSRF** (Weighted Subspace Random Forest)

In (Xu et al., 2012) is suggested a new random forest, namely, the WSRF model, which involves weighting the input variables and afterward opting for the variables that ensure each subspace always includes informative attributes. The WSRF model is implemented as multi-thread processes. This algorithm categorizes very high-dimensional data and sparse data with random forests made using small subspaces. A new variable weighting manner is applied for the variable subspace choice rather than the traditional random variable sampling in the random forest model (Zhao et al., 2017). More details of the WSRF model are presented in (Xu et al., 2012; Zhao et al., 2017)

**5.4 NB** (Naïve Bayes**) Method**

**Naïve Bayes** (NB) classifier is a Bayes-based classifier method which is used to maximize posterior possibility in a Bayesian setting as improve classification probability in which no dependency subsists between variables and conditioning factors exist (Pham et al., 2017; Pham et al., 2017b; Soni et al., 2011). The core function of NB classifier comprises (Khosravi et al., 2019);

- Selecting relevant training dataset;
- Determining the posterior probability of each class;
- Estimating class level and calculating co-variance matrix;
- Construct a discriminant function of each class level.

In (Ali et al. 2020) was found that random forest (RF) classifier model was highly accurate while Naïve Bayes classifier model had lower performance because it needs independence of criteria which makes it relatively robust. In (Soria et al., 2011) is shown that the prediction capability of the Naïve Bayes classification is comparatively lower than other consider models. By using ensemble machine learning and geographically weighted regression in (Chakrabortty et al., 2020) is shown easily predicts of soil erosion susceptibility compared to a physical-based empirical or semi-empirical model. Moreover, is shown a possibility to modify data on different scales, keeping in view the variability of climatic conditions and soil erosion causal factors. In (Mosavi et al., 2020) for the prediction of the water erosion of soil susceptibility in the Nur-Rood watershed are used ML methods, the WSRF, Gaussian process with a radial basis function (RBF) kernel and NB. The critical factors which were selected are aspect, curvature, slope length, flow accumulation, rainfall erosivity factor, distance from the stream, drainage density, fault density, normalized difference vegetation index (NDVI), hydrologic soil group, soil texture, and lithology. The results show that WSRF model have an acceptable performance regarding producing soil erosion susceptibility maps where its accuracy 91% while Gaussian process with RBF kernel achieve accuracy = 88% displayed better performances than the NB with accuracy = 85% model.

**5.5 ImpelERO** (Integrated Model to Predict European Land)

In (De la Rosa et al., 1999) is constructed a hybrid model of expert DTs and ANN (named ImpelERO) developed as USLE type model following traditional land evaluation analysis and advanced empirical modelling techniques to evaluate the soil erosion process. The model is use to develop an approach physically valid for a large spatial unit and to predict the water erosion depending on soil type and its management. It includes land and management qualities and their associated land and management characteristics grouped in subqualities. The physical input variables correspond to soil data available in standard soil surveys and climate monthly data. The main components of ImpelERO are: runoff erosivity, relief hazard, soil erodibility, crop protection, tillage translocation and productivity influence. The soil variables considered are referred to soil infiltration, cracking and mulching effects and aggregate stability. An ANN was trained using the Correlation-cascade algorithm and applied to capture the interactions between the land and management qualities and one output: vulnerability index to soil erosion. This model provided a powerful tool for assessing soil erosion risk at field information level in southern Spain as offers excellent performance in modelling the complex soil erosion problem, and very good quantification and generalization capability for prediction. It also be used to predict the positive or negative consequences of agricultural and climate changes which can take place in the future. This model was partly constructed according to the criteria of the FAO-framework for land evaluation (FAO, 1976). In (Diaz-Pereira et al., 2002) was considered the study area located at Coria del Rio (farm "La Hampa" of the Instituto de Recursos Naturales y Agrobiología de Sevilla, in the Sevilla province, Spain. The ImpelERO model using spatialization tool, DEM, improve soil erosion prediction as well as predicting the effects of management on sediment yield. The model is also tested in (Shahbazi et al., 2010) where the study area covers 14 natural regions in the north-west of Iran, east Azerbaijan province, for soil erosion assessment. The obtained assessments of soil erosion show that vulnerability areas with high risk are classes V4 and V5.

An application of ImpelERO model in (Afshar et al., 2016) was applied from Mashhad–Chenaran plain, Khorasan-e-Razavi province, northeast Iran, to quantify the soil erosion features including soil vulnerability index, soil loss rate, erosion risk class, and soil depth reduction by sugar beet cultivation under conventional and conservational management practices. By using of the ImpelERO model the soil erosion is simulated for four time horizons: current, 2020, 2050 and 2100. Soil erosion parameters were calculated on the basis of the ImpelERO model developed as a USLE model following traditional land evaluation analysis and advanced empirical modeling techniques and using n ANN approach.

Summary, the ImpelERO method is convenient to predict of agricultural soil erosion to investigate the effects of soil loss on crop yield in semi-arid region as well as to optimize the use of soil and prevent the increase in soil erosion and loss, with respect to which we can identify vulnerable segments of soil suitable for cultivation.

**5.6 SSAO-MARS** (Social Spider Algorithm Optimized the Multivariate Adaptive Regression Splines) **Method**

Since soil erosion is a complex process of analysis of various factors, is needed to be consider as highly variable depending on local climate, soil, topography, land-use, and land management practice. The empirical models, e.g. USLE, RUSLE, WEPP, and SWAT, generally rely on conventional data fitting methods they can hardly deal with the nonlinearity and multivariate which limited their applicability. Moreover, due to a fact that is needed to be calculated the location-dependent parameters these models suffer from low predictive accuracy.

MARS model was first introduced in (Friedman, 1991) as a methodology for nonparametrically estimating (and interpreting) general functions of a high-dimensional argument given (usually noisy) data. Its basic underlying assumption is that the function to be estimated is locally relatively smooth where smoothness is adaptively defined depending on the local characteristics of the function. There is developed a set of FORTRAN subroutines in (Friedman and Roosen, 2005), It is also is available as Matlab/Octave toolbox 'ARESLab' in (Jekabsons, 2016) with a major difference that the model building is not accelerated using the "fast least-squares update technique" (Friedman, 1991) which leads to the speed of the algorithm execution and the code might be slow for large data sets. Needed to note that ARESLab' does not automatically handle missing data or categorical input variables with more than two categories.

MARS is a non-parametric technique for dealing with supervising learning tasks. It incorporates a wide variety of predictor types in a natural manner, including both continuous and categorical variables. It has the ability to handle nested variables, and can adjust for missing values without discarding data. The model provides a way to fit a highly general regression model while avoiding overfitting as an extension of linear models that can recognize and mathematically present interaction between influencing attributes without any assumptions of distribution form. Generally, the construction phase of a MARS model involves of adding and removing basic functions and this method works efficiently for a certain sub-dataset. MARS model is consists from two phases (Friedman and Roosen, 2005),

- *The forward phase* – The forward part is an iterative (recursive) procedure as main goals is to reducing mean square error (MSE); each iteration simultaneously constructs an expanded list of basis functions to be considered and then decides which ones to enter at that step. Each an iteration adds two new basis functions to the current model.
- *The backward phase* – autonomously exclude basic functions that have less contribution to the final MARS model.

- A final appropriately sized basis function set is then selected through a backward stepwise variable subset selection procedure using the basis functions produced by the forward algorithm as candidate 'variables'.

An infinite combination of the hyper-parameters influence the learning phase and the predictive capability in machine learning and leads to a challenging task because the penalty coefficient can be searched in a continuous domain, example soil erosion case. Then a swarm intelligence based metaheuristic optimizer comes to help – the new social animal foraging strategy model to solve optimization problems algorithm – so called social spider algorithm (SSA) described in (Yu et al., 2015). Only with following 3 parameters evaluated a series of widely used benchmark functions has superior performance compared with other state-of-the-art metaheuristics:
- The rate of vibration attenuation when propagating over the spider web.
- The probability of the spiders changing their dimension mask in the random walk step.
- The probability of each value in a dimension mask to be one.

Therefore SSA has the potential to be applied to solve combinatorial problems as is included in other algorithms.

This is consider in (Vu et al., 2020) where machine learning based method for predicting soil erosion susceptibility in tropical hilly regions, Son La province (Northwest Vietnam) with data for (2009–2011), is proposed. It is Social Spider Algorithm (SSA) optimized the Multivariate Adaptive Regression Splines (MARS) method – called as SSAO-MARS. The algorithm SSA metaheuristic automatically fine-tunes hyper-parameters in MARS. The core is the MARS model strongly depends on the setting of the two hyper-parameters: the maximum number of basis functions and the penalty coefficient. It divided the learning space of soil erosion conditioning factors into sub-domains. The data in each sub-domain is then fitted by a linear mode as a classifier constructs a piecewise linear function to represent an adaptive global model. Additionally, to reduce overfitting, this machine learning utilizes a backward model refinement process. Thus, the MARS can be very potential for deal with the problem of interest. The experimental results show that the proposed hybrid approach of machine learning and swarm intelligence metaheuristic has achieved the most accurate prediction result of soil erosion susceptibility. One notable observation is that SSAO-MARS outperforms at all measurement indices both models – backpropagation artificial Neural Network (Montavon et al., 2012) and squares support vector machine (Samui, 2011). The SSAO-MARS model is only able to predict the soil erosion susceptibility and appropriate for predicting erosion susceptibility in areas with similar soil and climatic conditions.

**5.7 MPFPR** (Multi-Parameter Fuzzy Pattern Recognition)

It is a widely used method in the US development of PSIAC (Pacific Southwest Inter Agency Committee) (PSIAC, 1974). The PSIAC index uses a linear model to compute the value of the final erosion index. The assessment of soil erosion is a multiple parameter decision-making process with many difficult measure inexact or qualitative parameters. It has extended version with Multi-parameter fuzzy pattern recognition (MPFPR) considered in (Saadatpour et al., 2011). Based on fuzzy multi-parameter analysis and nonstructured decision making for multi-parameter systems, MPFPR method is consider for obtaining knowledge on the soil erosion and assessment of geological settings under different management strategies. It consists from two steps:

(a) The parameters set identification - surface topography, soil, climate, runoff etc;
(b) The standard matrix of soil property identification through the MPFPR decision making.

The idea of using fuzzy logic in erosion mapping is to consider the spatial objects on a map as members of a set. The results from the fuzzy method more efficiently reflected the fuzzy nature of the soil erosion and the effect of the parameters. The method is easily programmed and can effectively use the methodology to evaluate the soil erosion of catchments and solving of multi-parameters decision-making problems. It enables decision makers to evaluate the relative priorities of conserving water and soil projects, mitigation plan which should be implemented to control both the on-site and off-site impacts of erosion and land use activities based on a set of preferences, parameters, and indicators for the area.

Based of the efficiency of the fuzzy method in (Yang et al., 2014) the study area is Chao-tian Town, an area of Karst Mountains and hills in Guilin, northeast part of Guangxi Zhuang Autonomous Region, China is considered. The results show that fuzzy modeling of soil erosion risk indications is more clearly and better understand the different grades. The fuzzy method within GIS to crop-land suitability analysis and assess is also applied in (Nisar et al., 2000) by nine parameters (eight of soil and one of topography). The fuzzy membership classification allows consideration of partial memberships to obviate the limitations of classical classification methods while GIS consider the spatial variability of relevant terrain and other parameters. The advantage of the approach is delineating areas of various suitability ratings to a given crop with more accurately. In (Akyurek and Okalp, 2006) a theory of fuzzy sets and fuzzy logic algebra is used to compared and predict with USLE model for the soil erosion hazard. The fuzzification of the landscape elements uses Fuzzy Semantic Import modeling approach, *FuzzyCell* which is generic, and enables decision makers to express their constraints and imprecise concepts that are used with GIS through the use of natural language interfaces. In comparing "*crisp*" USLE versus "*fuzzy*" modeling, the fuzzy approach is successful at locating and differentiating areas of soil erosion with minimum data input. The results compared with traditional USLE model, show that traditional USLE overestimates the areas prone to low level erosion risks and it overestimates the areas prone to high level erosion risk, while the fuzzy logic approach is very useful spatial decision making.

## 6. DISCUSSIONS AND CONCLUSIONS

A preview of most uses and useful erosion models were considered. Uses of soil loss models suppose strong attention into reliability, predictive accuracy, and estimates of a model. Additionally, one must be careful considered the accurate of estimations of soil loss in order to implement conservation measures. For example, WEPP, SWAT, MMF, DYRIM, Si′s and Yang's model may be suitable for assessing the impacts of climate change, land-use shifts and land management. Since Si′s model is spatially-lumped, it is not able processing of the spatial pattern of erosional response over a large area. Because SWAT and WEPP take reservoirs into account and Tian's model – the sediment trapped by check-dams, they are useful for examining the impact of check-dams. Erosion control measures should always be adjusted to local conditions and simulation studies can help to identify the most optimal combination of measures.

One of the major challenges in soil erosion modelling is the mismatch between the small spatial and temporal scales of data collection and model conceptualization, and the large spatial and temporal scales. In erosion models main errors come from incompatibilities between model scales, scale of input parameter data, and the intended scale of the model outputs. It is clear that the variability, nonlinearity and the interacting nature of erosion and deposition processes over various scales significantly influence the mechanics of surface runoff generation and soil erosion.

An important problem is applications of the contemporary erosion models which used input data with resolution 1000m pixels. This difficulty cam be solved as assume data to be homogeneous for significantly larger areas than the 1000m resolution of pixels, such as Soil Mapping Units (SMUs).

The biggest problem with erosion modelling is the difficulty of validating the estimates produced. At the regional and larger scale, virtually no reliable data exist for comparing estimates with actual soil losses.

Controls on surface runoff generation and the resulting soil erosion are instantaneous phenomena and modelling them ideally requires fine time resolution data. There are methods which translating erosion models across spatial scales and diminishing the scaling effects as a great improvement in modelling results. Therefore, a key problem in choosing a model is the time scale at which the erosion processes will be predicted. Fortunately, today there are many and different models about estimate and prediction have its own characteristics and application scopes. Such that as USLE, RUSLE, EPIC, and the Geometric Mean Diameter based (Dg) model (Römkens et al., 1988). These were developed in the United States based on American soil-erosion databases, however, which may render them entirely ineffective for soils at other geographical locations. Others preview models are Water Erosion Prediction Project (WEPP), Soil Erosion Model for Mediterranean Regions (SEMMED), ANSWERS, LISEM, European Soil Erosion Model (EUROSEM), SWAT, Simulator for Water Resources in Rural Basins (SWRRB), Agricultural Non-point Source pollution model (AGNPS), etc. were used in regional scale assessment.

In conclusion, reduce of an erosion demands decrease of the deforestation, the conversion of the grassland to cropland, reduced tillage, plant residues, terracing in sloppy areas and contour farming. It fully corresponds with the current EU Common Agricultural Policy (CAP 2021–2027) and the EU Green Deal (European Commission CAP 2021–2027, 2018). To obtain better results should be pay of attention of the study areas or the country about an appropriate the choice of the soil erosion model to use.

The future the developments will be directed at improving and balancing the resolution of the input data against the convenience for model implementation as well as improving in soil erosion (process-based) models.

**ABBREVIATIONS**

AGNPS – Agricultural Non-Point Source Pollution model
AI – Artificial Intelligence
AISLUS – All India Soil and Land Use Survey
ALOS – Advanced Land Observing Satellite
ANFIS – Adaptive Neuro-Fuzzy Inference System
ANN – Artificial neural networks
ANSWERS – Aerial Non-Point Source Watershed Environment Response Simulation
CLSE – Chinese Soil Loss Equation
CNN – Convolutional neural network
CREAMS – Chemicals, Runoff and Erosion from Agricultural Management
CORINE – Coordinated Information on the Environment
DEM – Digital Elevation Model
DP – Deep Learning
DNN – Deep Neural Networks
DYRIM – Digital Yellow River Model
EI – Erosivity Index
EPIC – Erosion Productivity Impact Calculator
EPM – Erosion Potential Method
FLDA – Fisher's Linear Discriminant Analysis
GIS – Geographic Information System
GEIM – Gully Erosion Inventory Map
GPU – Graphical Processing Units
HAND – Height Above the Nearest Drainage
HI – Human Intelligence
ImpelERO – Integrated Model to Predict European Land
LAI – Leaf Area Index
LISEM – Limburg Soil Erosion Model
LDN – Land Degradation Neutrality
LMT – Logistic Model Tree
LR – Logistic Regression
LSP – Landslide Susceptibility Prediction
MARS – Multivariate Adaptive Regression Splines
ME – Model Efficiency
MESALES – Modèle d'Evaluation Spatiale de l'ALéa Erosion des Sols
ML – Machine Learning
MLP – Multilayer perceptron
MLR – Multi Linear Regression
MPFPR – Multi-Parameter Fuzzy Pattern Recognition
NBTree – Naïve Bayes Tree
NDVI – Normalized Difference Vegetation Index
NIR – Red and Near-Infra-Red
PESERA – Pan-European Soil Erosion Risk Assessment
QSWAT – Quantum Soil and Water Assessment Tool
RBF – Radial Basis Function

RF – Random Forest
RUSLE – Revised Universal Soil Loss Equation
SDG – Sustainable Development Goals
SDR – Sediment Delivery Ratio
SEMMED – Soil Erosion Model for Mediterranean Regions
SMUs – Soil Mapping Units
SONN – Self-Organization Neural Network
SSA – Social Spider Algorithm
SSAO-MARS – Social Spider Algorithm Optimized the Multivariate Adaptive Regression Splines
ST – Soil Taxonomy
SVM – Support Vector Machine
SVR – Support Vector Regression
SYI – Sediment Yield Index
SWAT – Soil and Water Assessment Tool
SWRC – Soil Water Retention Curve
SWRRB – Simulator for Water Resources in Rural Basins
USLE – Universal Soil Loss Equation
USPED – Unit Stream Power Erosion Deposition
WATEM/SEDEM – Water and Tillage Erosion Model/Sediment Delivery Model
WEPP – Water Erosion Prediction Project
WEPS – Wind Erosion Prediction System
WEQ – Wind Erosion Equation
WRB – World Reference Base
WSRF – Weighted Subspace Random Forest

# REFEENCES

Huang, F., Chen, J., Du, Z., Yao, C., Huang, J., Jiang, Q., Chang, Z., and Li, S. 2020. Landslide Susceptibility Prediction Considering Regional Soil Erosion Based on Machine-Learning Models, *International Journal of Geo-Information*, vol. 9, no. 6, article 377.

Gao, H., Hrachowitz, M., Fenicia, F., Gharari, S., and Savenije, H.H.G. 2014. Testing the realism of a topography-driven model (FLEX-Topo) in the nested catchments of the Upper Heihe, China, *Hydrology and Earth System Sciences*, vol. 18, no. 5, 1895–1915.

Grimm, M., Jones, R. and Montanarella, L. 2002. Soil Erosion Risk in Europe, https://esdac.jrc.ec.europa.eu/content/soil-erosion-risk-europe

Guo, D., Yu, B., Fu, X., and Li, T. 2015. Improved Hillslope Erosion Module for the Digital Yellow River Model, *Journal of Hydrologic Engineering*, vol. 20, no. 6.

G2. https://esdac.jrc.ec.europa.eu/themes/g2-model, (date accessed 15 November 2020)

FAO: A frarnework for land evaluation. 1976. Soils Bull., 32, Rome. FAO.

Feng, X., Wang, Y., Chen, L., Fu, B., and Bai, G. 2010. Modeling soil erosion and its response to land-use change in hilly catchments of the Chinese Loess Plateau, *Geomorphology*, vol. 118, 239–248.

Flanagan, D. C., Gilley, J. E., and Franti, T. G., 2007. Water erosion prediction project (WEPP): development history, model capabilities and future enhancements. In: *Transactions of the American Society of Agricultural and Biological Engineers*, vol. 50, 1603–1612.

Flanagan, D., Frankenberger, J., and Ascough, J. 2012. WEPP: model use, calibration, and validation, *Trans. Am. Soc. Agric. Biol. Eng*., vol. 55, 1463–1477.

Friedman, J.H. 1991. Multivariate adaptive regression splines, Ann. Statist., vol. 19, 1–67.

Friedman, J.H. and Roosen, C. B. 1995. An introduction to multivariate adaptive regression splines, *Stat. Methods Med. Res*., 4, 197–217.

Fu, B.J., Zhao, W.W., Chen, L.D., Zhang, Q.J., Lü, Y.H., Gulinck, H., and Poesen, J. 2005. Assessment of soil erosion at large watershed scale using RUSLE and GIS: a case study in the Loess Plateau of China. *Land Degradation & Development*, vol. 16, 73–85.

Funk, R., Skidmore, E.L., and Hagen, L.J. 2004. Comparison of wind erosion measurements in Ger-many with simulated soil losses by WEPS. Environ. Model. Softw., vol. 19, no. 2, 177–183.

Gajbhiye, S., Mishra, S.K., Pandey, A. 2015. Simplified sediment yield index model incorporating parameter curve number, Arabian Journal of Geosciences, vol. 8, 1993–2004.

GeoGnostics, Erosion3D — Simulierte Prozesse, http://geognostics.de/e3d_prozesse.html (date accessed 15. Nov. 2020)

Grunwald, S. and Norton, L.D. 1999. An AGNPS-based runoff and sediment yield model for two small watersheds in Germany, Transactions of ASAE, vol. 42, no. 6, 1723–1731.

Kamilaris, A. and Prenafeta-Boldú, F.X. 2018. Deep learning in agriculture: A survey, *Comput. Electron*. Agr., vol. 147, 70–90.

Karaburun, A. 2010. Estimation of C factor for soil erosion modeling using NDVI in Buyukcekmece watershed, *Ozean J. Appl. Sci*., vol. 3, 77–85.

Karamesouti, M., Petropoulos, G.P., Papanikolaou, I.D., Kairis, O., and Kosmas, K. 2016. Erosion rate predictions from PESERA and RUSLE at a Mediterranean site before and after a wildfire: comparison & implications, *Geoderma*, vol. 261, 44–58.

Karydas, C.G., Panagos, P., and Gitas, I.Z. 2012. A classification of water erosion models according to their geospatial characteristics, *International Journal of Digital Earth*, vol. 7, no. 3, 229–250.
46